\documentclass[5p,authoryear,twocolumn,lefttitle]{elsarticle}

\usepackage{amssymb}
\usepackage{amsmath}
\usepackage{fancyhdr}
\usepackage{algorithm2e}
\usepackage{enumitem}
\RestyleAlgo{ruled}

\usepackage{graphicx}
\usepackage[table,xcdraw]{xcolor}
\usepackage{booktabs}
\usepackage{multirow}
\usepackage{graphicx}    
\usepackage{float}       
\usepackage{placeins}
\usepackage{caption}
\usepackage{subcaption}
\usepackage{tikz}
\usetikzlibrary{decorations.pathreplacing}  
\usepackage{siunitx}    

\definecolor{CAD7E5}{HTML}{CAD7E5}

\usepackage[T1]{fontenc}
\usepackage[utf8]{inputenc}
\newcolumntype{Y}{>{\raggedright\arraybackslash}X}
\newcommand{\code}[1]{\texttt{\footnotesize #1}}
\newcolumntype{G}{>{\raggedright\arraybackslash}p{0.35\textwidth}}
\newcolumntype{C}{>{\centering\arraybackslash}p{0.12\textwidth}}                         
\newcolumntype{P}{>{\raggedright\arraybackslash}p{0.18\textwidth}}                       

\definecolor{riskred}{HTML}{E45756}
\definecolor{safelight}{HTML}{E8F8F2}
\definecolor{rowshade}{HTML}{F7F9FC}

\usepackage[
    colorlinks=true,
    linkcolor=red,    
    citecolor=red,   
    urlcolor=blue    
]{hyperref}

\usepackage{booktabs,array} 
\usepackage{cleveref}
\usepackage{colortbl}
\journal{Journal of Safety Research}

\begin{document}

\begin{frontmatter}

\title{A new machine learning framework for occupational accidents forecasting with safety inspections integration}

\author[inst1,inst2]{Aho Yapi\corref{mycorrespondingauthor}}
\cortext[mycorrespondingauthor]{Corresponding author at : Campus universitaire des Cézeaux TSA 60026 - CS 60026 3, Place Vasarely 63178 AUBIERE}
\ead{A-Aymar.YAPI@doctorant.uca.fr}
\author[inst1,inst3]{Pierre Latouche}
\author[inst1]{Arnaud Guillin}
\author[inst2]{Yan Bailly}

\affiliation[inst1]{organization={Laboratoire de Mathématique Blaise Pascal UMR 6620 CNRS, Université Clermont Auvergne},%
            addressline={place Vasarely, 63178}, 
            city={Aubière},
            postcode={63170}, 
            country={France}}

\affiliation[inst2]{organization={LYF SAS},%
            addressline={27 rue Raynaud}, 
            city={Clermont-Ferrand}, 
            postcode={63000}, 
            country={France}}
            
\affiliation[inst3]{organization={Institut Universitaire de France (IUF)},%
            city={Paris}, 
            country={France}}

\begin{abstract}
\textit{Introduction}: Reducing the number of occupational accidents remains a major challenge for companies, as these events lead to significant human harm and financial losses. Although many organizations have implemented safety programs and made continuous efforts to improve their prevention strategies, these measures often remain insufficient to proactively and dynamically anticipate risks. In particular, safety inspections are still largely underexploited, and their integration into continuously updated predictive models has received little attention. \textit{Methods:} we propose a model-agnostic framework for short-term occupational accident forecasting that leverages safety inspections and models accident occurrences as binary time series. The approach generates daily predictions, which are then aggregated into weekly safety assessments for better decision making. To ensure the reliability and operational applicability of the forecasts, we apply a sliding-window cross-validation procedure specifically designed for time series data, combined with an evaluation based on aggregated period-level metrics. Several machine learning algorithms, including logistic regression, tree-based models, and neural networks, are trained and systematically compared within this framework. \textit{Results:} across all tested algorithms, the proposed framework reliably identifies upcoming high-risk periods and delivers robust period-level performance, demonstrating that converting safety inspections into binary time series yields actionable, short-term risk signals. \textit{Conclusions and Practical Applications:} the proposed methodology converts routine safety inspection data into clear weekly and daily risk scores, detecting the periods when accidents are most likely to occur. Decision-makers can integrate these scores into their planning tools to classify inspection priorities, schedule targeted interventions, and funnel resources to the sites or shifts classified as highest risk, stepping in before incidents occur and getting the greatest return on safety investments.
\end{abstract}

\newpageafter{abstract}

\begin{keyword}
Occupational accident prevention \sep Proactive safety management \sep Binary time series \sep Machine learning \sep Sliding-window cross-validation \sep Safety inspections
\end{keyword}

\end{frontmatter}


\section{Introduction}
\label{sec:intro}

The international labour organization (ILO) estimates that nearly~300,000 people die each year due to occupational accidents \citep{ILO2023}. In France, the national health insurance  recorded over 600 000 occupational accidents in 2023, nearly 700 of them fatal and this level has remained essentially flat for over a decade \citep{Amelireports}. This observation highlights the limitations of current strategies and the urgent need for new approaches to sustainably reduce both the frequency and severity of occupational accidents.

Since Heinrich’s pioneering work \citep{heinrich1931} and his domino theory, the understanding of occupational accidents has evolved considerably. These events are no longer seen as isolated or random, but rather as the outcome of a chain of contributing factors with complex interactions. Several theoretical models have emerged, such as Surry's sequence of events model \citep{surry1969}, Reason's Swiss cheese model \citep{reason1990}, and the cause tree method developed by the French INRS institute \citep{inrs1970}. While these models have helped structure accident investigation processes, they remain primarily retrospective, lack predictive capability, and often fail to capture the temporal dynamics of risk in complex environments \citep{Qureshi2007, Larouzee2020}.

With the growing adoption of machine~learning, new proactive strategies have
been introduced in sectors such as construction, mining, agriculture, and
services. Numerous studies demonstrate the potential of these techniques for
incident prediction \citep{SuarezSanchez2011,Rivas2011,Wang2018},
risk assessment \citep{Palei2009,Weng2011,Leu2013},
injury severity classification \citep{Chang2013,Esmaeili2015,Tixier2016},
and risk\nobreakdash-factor identification \citep{Cheng2012,Amiri2016}.
However, most of these approaches rely on lagging indicator data
collected after an accident, which limits their usefulness for anticipating
risks and taking preventive action in advance.
To address this limitation, the concept of safety leading indicators (SLIs) has gained momentum. In contrast to lagging indicators, SLIs allow for the early detection of weak signals based on proactive field data \citep{Reiman2012, Hinze2013}. Numerous studies emphasize their usefulness for constructing predictive models \citep{Grabowski2007, Poh2018, Gondia2023}. However, most of the existing frameworks using SLIs suffer from two major limitations: (1) they are rarely updated continuously, limiting their ability to adapt to evolving operational contexts; and (2) they often fail to explicitly capture temporal dependencies and to integrate recent information into the prediction process. 

Time series models have also been applied to accident forecasting \citep{carnero2010modelling,koc2022}, but they typically operate at national or regional levels, over long time spans. Other works focus on building early warning systems based on composite indicators \citep{LI201619,nazaripour2018safety}, yet these systems require domain-expert thresholds and domain-specific calibration, limiting their operational flexibility. 
Despite these modeling advances, traditional methods remain predominant in occupational safety practice most notably the Fine–Kinney risk score \citep{Fine1971,KinneyWiruth1976} and generic risk matrices \citep{ISO31010}. These approaches assign ordinal hazard indices from expert judgment and periodically aggregated indicators to prioritize controls. While simple and widely adopted, they depend on subjective scales and fixed thresholds and may compress quantitatively different risks \citep{Cox2008}.
In this paper, we present a generic framework for short-term forecasting of occupational accidents at the company and department levels. We model daily accident occurrence as a binary time series \citep{KedemFokianos2002,FokianosKedem2003}. Unlike prior prediction and risk-scoring approaches, we explicitly learn day-to-day dynamics and cast the task as multi-output time-series classification. Conditioned on strictly ex-ante, continuously updated indicators from safety inspections, the model jointly predicts, for each site or department, the probability of at least one accident on each of the next $H$ days. We implement direct multi-horizon (MIMO) and direct-recursive (DirRec) strategies to produce probabilities for each forecast horizon. We convert these calibrated day-level probabilities into an interpretable weekly status: a day is flagged at risk when its probability exceeds a calibrated threshold, and a week is labeled \textit{risky} if it contains at least one such day. This design leverages daily temporal dependence rather than aggregated counts, yields continuously updated risk maps across multiple horizons for planning instead of a single static score, and enables transparent, decision-oriented evaluation at every horizon. We assess performance with metrics suited to imbalanced classification \citep{Luque2019} and validate robustness using sliding-window cross-validation tailored to time series \citep{Tashman2000,bergmeir2012crossval,hyndman2018forecasting}, which replicates the operational rollout procedure.
For illustration, Table~\ref{tab:two_weeks} shows the output of our approach applied to a specific department within the company under study, over a two-week period. The daily threshold for binarizing accident risk is set based on a calibration process and fixed here at $0.6$. In week 22, one day exceed this threshold, resulting in the classification of the entire week as \textit{risky}. In contrast, in week 23, no day probability crosses the threshold, so the week is classified as \textit{safe}. This example highlights how the approach can be used at the departmental level to provide timely and actionable insights for occupational risk management.
\begin{table*}[ht]
    \centering
    \renewcommand{\arraystretch}{1.25}
    \setlength{\tabcolsep}{8pt}
    \resizebox{\textwidth}{!}{%
    \begin{tabular}{@{}l*{7}{c}@{\hspace{1.2em}}*{7}{c}@{}}
        \toprule
        & \multicolumn{7}{c}{\textbf{Week 22}} & \multicolumn{7}{c}{\textbf{Week 23}} \\
        \cmidrule(lr){2-8}\cmidrule(lr){9-15}
        \textbf{Day} & \textbf{Mon} & \textbf{Tue} & \textbf{Wed} & \textbf{Thu} & \textbf{Fri} & \textbf{Sat} & \textbf{Sun}
                     & \textbf{Mon} & \textbf{Tue} & \textbf{Wed} & \textbf{Thu} & \textbf{Fri} & \textbf{Sat} & \textbf{Sun} \\
        \midrule
        \textbf{Daily probability} 
        & \cellcolor{safelight}0
        & \cellcolor{safelight}0
        & \cellcolor{safelight}0
        & \cellcolor{riskred!30}\textbf{0.65}
        & \cellcolor{safelight}0
        & \cellcolor{safelight}\textbf{0.39}
        & \cellcolor{safelight}0
        & \cellcolor{safelight}0
        & \cellcolor{safelight}0
        & \cellcolor{safelight}0
        & \cellcolor{safelight}0
        & \cellcolor{safelight}0
        & \cellcolor{safelight}0
        & \cellcolor{safelight}0 \\
        \addlinespace[2pt]
        \textbf{Forecast (0/1)}
        & \textbf{1} & \textbf{1} & 0 & 0 & 0 & 0 & 0
        & 0 & 0 & 0 & 0 & 0 & 0 & 0 \\
        \addlinespace[4pt]
        \rowcolor{rowshade}
        \textbf{Predicted accidents}
        & \multicolumn{7}{c}{\textbf{2}} & \multicolumn{7}{c}{\textbf{0}} \\
        \rowcolor{rowshade}
        \textbf{Actual accidents}
        & \multicolumn{7}{c}{1} & \multicolumn{7}{c}{0} \\
        \rowcolor{rowshade}
        \textbf{Bias (Pred. – Actual)}
        &\multicolumn{7}{c}{\textbf{1}} & \multicolumn{7}{c}{\textbf{0}} \\
        \addlinespace[2pt]
        \textbf{Period status}
        & \multicolumn{7}{c}{\cellcolor{riskred!35}\textit{Risky}}
        & \multicolumn{7}{c}{\cellcolor{safelight}\textit{Safe}} \\
        \bottomrule
    \end{tabular}%
    }
    \vspace{2mm}
   \caption{Output of the proposed approach for a single department over two consecutive weeks. Weeks are flagged as \textit{risky} or \textit{safe} by comparing each day’s accident probability with a calibrated threshold of 0.6; week 22 is labelled \textit{risky} because two days exceed the threshold, whereas week 23 remains \textit{safe} since no day does.}
    \label{tab:two_weeks}
\end{table*}

\section{Literature review} 
\label{sec:related}

\subsection{Leading vs. lagging safety indicators}

An indicator is a qualitative or quantitative measure used to assess or monitor the evolution of a situation, phenomenon, or activity. In the field of data-driven occupational safety management, two main types of indicators are typically distinguished \citep{ Grabowski2007, Hopkins2009}: lagging indicators and leading indicators. Lagging indicators such as  accident or incident rates, compensation costs and number of injuries resulting in time off work \citep{Choudhry2007, Hinze2013, Jazayeri2017} reflect the consequences of accidents that have already occurred. In other words, they are updated only after an accident happens. Several authors \citep{Grabowski2007, Mengolini2008} argue that such indicators do not provide sufficiently useful information to prevent future accidents. According to \citet{Lindsay1992}, a low number of reported accidents even over several years does not necessarily mean that risks are under control or that other incidents will not occur. Despite their limitations, these indicators remain widely used because they are easy to quantify and identify \citep{Lingard2013, Almost2018} and allow organizations to benchmark against one another. \citep{Elsebaei_2020}.

In contrast, leading indicators provide early warning signs of accidents and adopt a more proactive approach, aiming to detect and act before incidents occur \citep{Mearns2009, Eaton2013}. Examples include near-miss report, safety talks, and safety inspections \citep{Falahati2020}. Field-level feedbacks can also be added to this list, as they enable quick and spontaneous collection of real-world operational data, helping to manage weak signals in real time. In the construction sector, \citet{Hinze2013} highlight the importance of these indicators and differentiates between \emph{passive} leading indicators (e.g., number of employees trained or presence of a prevention plan) and \emph{active} leading indicators (e.g., the percentage of safety meetings attended by supervisors). The latter reflects more accurately the dynamic reality of prevention efforts.

An illustrative analogy to differentiate leading and lagging indicators is that of driving a car: the dashboard (speed, fuel level, GPS) corresponds to leading indicators, providing real-time information to anticipate risks, whereas the odometer (distance traveled) is a lagging indicator, offering retrospective data about what has already occurred.

Despite their promise, leading indicators remain difficult to adopt widely, partly due to the diversity of work environments: an effective indicator in construction may not apply in agriculture or maritime industries \citep{Hinze2013, Xu2021}. Furthermore, the subjectivity associated with some indicators such as the assessment of the severity of a hazardous situation can distort the actual perception of system performance or activity \citep{Grabowski2007}. Sometimes, the boundary between leading and lagging indicators is blurry, and some indicators are poorly defined or misaligned with their intended objectives. It is therefore crucial to distinguish between process safety hazards risks inherent to the operation of the system (e.g., explosions or toxic spills) and personal safety hazards  which are more related to individual accidents such as falls, crushes, or electrocutions \citep{Hopkins2009}. Common lagging indicators like accident rates are often focused on personal safety and fail to capture process-related risks effectively. Similarly, some leading indicators (e.g., audit frequency) can remain too generic if they do not account for the specific processes of the company, thus failing to assess the actual quality of process safety.

To be truly effective, indicators must be clearly defined with respect to their scope of application: it must be stated upfront whether they concern process safety or personal safety, in order to properly evaluate the relevant prevention and risk management efforts. Although organizations collect a wealth of proactive data, they often lack the motivation or tools to make use of them, and it is often difficult to demonstrate the predictive power of such data. In this context, machine learning approaches can assist in identifying and even designing new leading indicators \citep{Poh2018, Gondia2023}, paving the way for more targeted and effective prevention.

\subsection{Traditional safety scoring systems and manual risk assessments}
\label{subsec:trad_scoring}
Traditional safety scoring systems and manual risk assessments remain widely used in industry because they are simple, familiar, and inexpensive. In the Fine--Kinney method, risk is rated by multiplying ordinal scores for probability, exposure, and consequence to obtain a priority index for action \citep{Fine1971,KinneyWiruth1976}. Generic risk matrices, similarly map qualitative or semi-quantitative ratings to risk bands that guide decision making \citep{ISO31010}. These tools rely on expert judgment and indicators aggregated at periodic intervals (e.g., monthly audit findings or incident counts), which makes them practical for audits and routine reviews. However, they also have well documented limitations: scales are subjective and ordinal, fixed thresholds can distort prioritization, and combining categories may compress risks that differ in magnitude; more fundamentally, such schemes do not model day-to-day dynamics or provide horizon-specific accident probabilities, which are increasingly needed for proactive planning \citep{Cox2008}. In this study, we view these traditional methods as complementary: they remain useful for screening and communication, while our time series framework transforms leading indicators into daily, horizon-resolved probability forecasts that support short-term operational decisions.

\subsection{Predictive models for occupational accident prevention}

In recent years, the use of predictive models based on machine learning has become increasingly widespread in occupational safety, thanks to their ability to identify leading indicators \citep{Gondia2023, Poh2018} and extract various risk factors \citep{kang2019occupational_accidents, choi2020fatal_accidents}. Organizations collect vast amounts of data without always being able to detect the weak signals that would help initiate relevant preventive actions \citep{Mearns2009, Tixier2016}. To address this challenge, several algorithms have been deployed, including logistic regression, decision trees, random forests, boosting models, and neural networks \citep{kim2024deep}. These techniques are applied across many sectors, such as construction \citep{Tixier2016, Gondia2023, Poh2018}, maritime transport \citep{Kretschmann2020}, metallurgy \citep{Sarkar2020b}, and the service industry \citep{Matias2008}.

In the construction sector, \citet{Poh2018} compare various algorithms including logistic regression, decision trees, random forests, and SVMs to classify construction sites according to their safety level. The results show that random forests outperform the other models (see Table~8 in \citet{Poh2018}). Similarly, \citet{Gondia2023} use five predictors such as site environment, hazard exposure, human error, familiarity with the site, and current month to test algorithms such as naive Bayes, decision trees, random forests, SVMs, neural networks, and an ensemble model based on weighted voting. The ensemble approach yields better performance than any individual component \cite[see][Table~7]{Gondia2023}. The resulting prediction probabilities are used as leading indicators to assess site-level risk and enhance prevention efforts. \citet{Kretschmann2020} also explores accident forecasting,
introducing inspection-based indicators to anticipate
safety conditions on ships, and using random forests for prediction.

National databases have also been used to identify the workers most exposed to fatal accidents. For example, \citet{koc2023} analyzes 338,173 accidents in the Turkish construction sector using a combination of random forests, particle swarm optimization, and SHAP analysis, highlighting the importance of age, job position, experience, salary, and accident history. Similarly, \citet{choi2020fatal_accidents} leverage a large Korean dataset to predict fatality risks, comparing several models and confirming the superiority of random forests. These studies demonstrate that integrating national data and detailed worker-level indicators (e.g., age, role, seniority) enhances the ability to identify high-risk situations and key contributing factors.

Despite their effectiveness, these models still have some limitations. Many studies rely on monthly data granularity: they typically use only the previous month information, without accounting for weekly fluctuations or longer-term trends. As a result, sudden changes or minor incidents may go unnoticed between two monthly observations. This lack of continuous tracking prevents the model from capturing abrupt increases in risk, thus limiting the responsiveness of preventive measures. Additionally, the sequential nature of the data is often overlooked, which prevents from capturing both long-term dynamics and short-term variations, ultimately reducing the model ability to anticipate increasing risks.

\subsection{Time series modeling and occupational accidents}

A time series refers to a set of data collected at regular intervals, enabling the analysis of trends and the evolution of a phenomenon over time. In the context of occupational safety, such methods have primarily been applied at large scales over extended periods to uncover global trends, inform public policy, and compare the performances of companies in terms of accident prevention strategies \citep{carnero2010modelling, melchior2021forecasting}.

Numerous studies rely on classical statistical models to investigate workplace accidents. \citet{melchior2021forecasting} use various ARMA variants to estimate monthly mortality rates while \citet{carnero2010modelling} and \citet{verma2023forecasting} employ ARIMA and unobserved components models to forecast incident frequencies. \citet{nazaripour2018safety} and \citet{LI201619} propose global indices designed to anticipate risk. \citet{nazaripour2018safety} develop the customized predictive risk index (CPRI) using AR and MA models to assess safety performance in a steel plant, while \citet{LI201619} introduce an early warning system that combines multiple composite indices with a $GM(1,1)$ model. In these approaches, defining and interpreting thresholds requires substantial domain expertise to appropriately guide preventive actions.

Some studies have focused on leveraging machine learning models to forecast accident time series. \citet{koc2022} apply wavelet decomposition to handle data non-stationarity and then use several algorithms, including artificial neural networks (ANN), support vector regression (SVR), and multivariate adaptive regression splines (MARS), to predict the daily number of accidents over short-, medium-, and long-term horizons. Their study relies on 393,160 construction-related accidents reported in Turkey between 2012 and 2020 and shows that integrating wavelets significantly improves forecast accuracy.

Although these works explore a wide range of methods and application domains, several limitations remain. Many studies still rely on univariate time series focusing solely on the number of accidents or mortality rates, without incorporating covariates such as safety inspections that could provide deeper insight into risk factors. Moreover, to our knowledge, binary time series models explicitly addressing the question “Will an accident occur in the short term?” have not yet been explored.

\subsection{Deep learning for time series and occupational accident forecasting}

Despite growing interest, deep learning remains comparatively underused in the occupational-accident literature; in particular, applying sequence models to \emph{binary} day-level accident time series is still uncommon. Recent reviews indicate that accident prevention applications have focused mainly on computer vision and narrative text analysis rather than day-level forecasting of accident occurrence \citep{Liu2022_AutCon}. Nevertheless, there is accumulating evidence that modern neural approaches can add value in safety-related contexts. For example, \citet{kim2024deep} explored deep neural networks (DNN), long short-term memory (LSTM), and recurrent neural networks (RNN) to estimate fatality probabilities under natural hazards by combining geographical, climatic, and construction site covariates. After benchmarking 36 architectures, they found that an Adam-optimized DNN attains the highest accuracy \citep[Table~3]{kim2024deep}. Looking forward, several deep sequence-modeling families are directly applicable to day-level accident-risk forecasting. Thus temporal convolutional networks (TCN) use causal, dilated convolutions with residual connections to capture long effective memory with stable gradients and often challenge RNN or LSTM baselines on sequence benchmarks \citep{Bai2018_TCN}. Moreover, transformer-based time series models most notably the temporal fusion transformer (TFT) can produce multi-horizon forecasts alongside variable and horizon-level interpretability via gating, variable selection, and attention \citep{Lim2021TFT}.

\section{Methods}
\label{sec:methods}

\subsection{Forecasting accident risk via binary time series modeling}

We represent the daily occurrence of accidents using a binary time series $\{y_t\}_{t=1}^{T}$, where $y_t = 1$ if at least one accident occurs on day $t$ and $y_t = 0$ otherwise.  
Two predictor families are distinguished: (i) \emph{future calendar covariates} $s_t$ such as month or day of week, which are fully known for any future date; and (ii) \emph{dynamic inspection covariates} $c_t$ extracted from the most recent safety inspection report available at day $t$.

Our aim is to estimate, for each step $h\in\{1,\dots,H\}$, the probability that at least one accident will occur.

\[
  p_{t+h}
  \;=\;
 \mathbb{P} \!\bigl(y_{t+h}=1 \mid Y_{t}^{d_y}, C_{t}^{d_c},  s_{t+h}\bigr),
\]
with
\[
  Y_{t}^{d_y}
  \;=\;
  \bigl(
      y_{t},\dots,y_{t-d_y+1}
  \bigr), \qquad 
  C_{t}^{d_c}
  \;=\;
  \bigl(
    c_{t},\dots, c_{t-d_c+1}
  \bigr),
\]
where \(d_c,d_y\ge 1\) are the numbers of lagged days for the dynamic covariates
\( c_t\) and the binary outcomes \(y_t\), respectively, and
\( s_{t+h}\), the static calendar features for day \(t+h\).

Finally, each predicted probability \(\hat{p}_{t+h}\) is turned into a binary class using a threshold
$\tau \in [0, 1]$:
$$
\hat{y}_{t+h} = \mathbf{1}_{\{\hat{p}_{t+h} \ge \tau\}}.
$$

\subsection{Multi-step forecasting strategies and evaluated machine learning models}

Our aim  is to predict the sequence $\{y_{t+h}\}_{h=1}^{H}$, thus producing forecasts for multiple future time steps. Several strategies can be adopted \citep{Bontempi2013}, which are commonly categorised by the output dimensionality of the underlying model. Our framework is model-agnostic. We therefore evaluate both classical single-output learners (used with the Direct–Recursive strategy) and multiple-output deep models (MIMO or encoder–decoder). Table~\ref{tab:models_strategies_slim} summarizes the models and their multi-step mapping.

\subsubsection{Direct recursive strategy (DirRec)}

A single-output learner produces one step ahead at a time. To extend it to multi-step forecasting, we use the \textit{DirRec} (Direct-Recursive) strategy~\citep{Sorjamaa2006timeseries},
which combines direct and recursive methods. It trains a separate estimator
\(f_h(\cdot;\theta_h)\) for every horizon \(h = 1,\dots,H\). Except for the first, each estimator receives the forecasts produced at earlier horizons as additional inputs.
Accordingly, the one-step-ahead forecast is
\[
  \hat{p}_{t+1}
  \;=\;
  f_{1}\!\bigl(
      y_{t},\dots,y_{t-d_y+1},\;
      C_{t}^{d_c},\;
      s_{t+1};\theta_{1}
  \bigr).
\]
We recursively pass the preceding
predictions to the horizon-specific learner for \(h = 2,\dots,H\):
\[
  \hat{p}_{t+h}
  \;=\;
  f_{h}\!\bigl(
      \hat{p}_{t+h-1},\dots,\hat{p}_{t+1},\;
      y_{t},\dots,y_{t-d_y+1},\;
      C_{t}^{d_c},\;
      s_{t+h};\theta_{h}
  \bigr).
\]
DirRec limits error propagation compared with pure recursion,
because each horizon has its own parameters \(\theta_h\), yet still
captures inter-horizon dependencies overlooked by the fully direct
strategy. The trade-off is increased training time and memory (one model
per horizon) together with a residual risk of bias accumulation through
the reused forecasts.

\subsubsection{Multiple-input multiple-output (MIMO)}

A \emph{multiple-output} learner returns an $H$-dimensional prediction vector in a single forward pass, eliminating the need for iterative generation of successive horizons. By producing all future points simultaneously, these learners can exploit cross-horizon dependencies that single-output strategies must ignore or approximate.

MIMO is the standard strategy for multiple-output forecasting \citep{BenTaieb}.
A single estimator $f(\cdot;\theta)$ simultaneously returns the complete forecast vector for entire horizon $H$:
$$
  (\hat{p}_{t+1},\dots,\hat{p}_{t+H})
  \;=\;
  f\!\bigl(
      y_{t},\dots,y_{t-d_y+1},\;
      C_{t}^{d_c},\;
      s_{t+1:t+H};\theta
  \bigr).
$$
Because every horizon is predicted directly from observed data, MIMO
avoids the error accumulation associated with recursive schemes and, unlike the fully direct approach, explicitly captures cross-horizon dependencies within a single shared parameter set \(\theta\).

\subsubsection{Autoregressive Seq2Seq architecture}
\label{sec:seq2seq}

We also consider an encoder--decoder (\emph{Seq2Seq}) architecture with an \emph{autoregressive} decoder \citep{Sutskever2014}.
An encoder maps the recent history $\big(Y_t^{d_y}, C_t^{d_c}\big)$ into a latent state
\[
  \mathbf{z}_t \;=\; \mathrm{enc}\!\bigl(Y_t^{d_y}, C_t^{d_c};\,\psi\bigr),
\]
where \(\psi\) are the encoder parameters.
and a decoder then produces accident probabilities step by step, each conditioned on the previous output and on the \emph{known} future calendar covariates:
\[
  \hat{p}_{t+h}
  \;=\;
  \sigma\!\Big(
     \mathrm{dec}\!\big(\mathbf{z}_t,\; u_{t+h-1},\; s_{t+h};\,\phi\big)
  \Big),\qquad h=1,\dots,H,
\]
with \(u_t = y_t\) (last observed outcome) for \(h=1\) and, for \(h \ge 2\), \(u_{t+h-1} = \hat{p}_{t+h-1}\). Moreover,
\(\sigma(\cdot)\) denotes the logistic function and \(\phi\) are the decoder parameters.
While MIMO predicts the entire vector $(\hat{p}_{t+1},\dots,\hat{p}_{t+H})$ in a single forward pass with a shared parameter set, an \emph{autoregressive} seq2seq decoder predicts one step at a time, each step using the previous output. This makes the forecast length flexible (the decoder can be unrolled to any $H$) and lets day-to-day effects carry over, but early errors may propagate to later steps. By contrast, MIMO avoids this roll-out effect but captures cross-horizon structure only implicitly through its shared representation. A non-autoregressive encoder--decoder can also produce all horizons in a single forward pass. it avoids roll-out error but does not condition on previous predictions.

\begin{table*}[ht]
\centering
\renewcommand{\arraystretch}{1.15}
\setlength{\tabcolsep}{6pt}
\resizebox{\textwidth}{!}{%
\begin{tabular}{l l l l p{0.28\textwidth}}
\toprule
\textbf{Family} & \textbf{Model} & \textbf{Output type} & \textbf{Multi-step strategy} & \textbf{References} \\
\midrule
\multirow{7}{*}{Machine learning}
 & Logistic regression & Single-output & DirRec & \citep{bishop2006pattern} \\
 & Linear discriminant analysis & Single-output & DirRec & \citep{bishop2006pattern} \\
 & Decision tree (CART) & Single-output & DirRec & \citep{breiman1984cart} \\
 & Random forest & Single-output & DirRec & \citep{breiman2001random} \\
 & HistGradient Boosting & Single-output & DirRec & \citep{friedman2001greedy,ke2017lightgbm} \\
 & XGBoost & Single-output & DirRec & \citep{xgboost,friedman2001greedy} \\
 & LightGBM  & Single-output & DirRec & \citep{ke2017lightgbm} \\
\midrule
\multirow{5}{*}{Deep learning}
 & Multilayer perceptron (MLP) & Vector of size $H$ & MIMO & \citep{bishop1995nn} \\
 & LSTM--MIMO & Vector of size $H$ & MIMO & \citep{HochreiterSchmidhuber1997,taieb2015review} \\
 & LSTM--Seq2Seq & One-step output & Seq2Seq (autoregressive) & \citep{Sutskever2014,cho2014seq2seq} \\
 & TCN & Vector of size $H$ & MIMO & \citep{bai2018tcn} \\
 & TFT & Vector of size $H$ & Seq2Seq (non-autoregressive) & \citep{Lim2021TFT} \\
\bottomrule
\end{tabular}
}
\caption{evaluated models, output types and multi-step strategies.}
\label{tab:models_strategies_slim}
\end{table*}

\subsection{Period-level risk assessment and evaluation metrics}

While accurately predicting the exact date of an accident would be ideal, it is rarely feasible. consequently, the prevailing objective is to evaluate whether a specified time interval is characterized by elevated risk. Aggregating data by week, for instance, helps smooth out daily fluctuations and emphasizes the overall occurrence of accidents within the period. This approach is analogous to intermittent demand problems in inventory management \citep{Croston1972,SyntetosBoylanCroston2005,wallstrom2010evaluation}, where the focus is placed on stock availability over a period rather than on precise daily tracking.

\begin{figure*}[t]
\centering
\resizebox{\textwidth}{!}{%
\begin{tikzpicture}[x=0.55cm,y=1cm,>=stealth]

\draw[->] (0,0) -- (30,0) node[below] {\small time $t$};

\foreach \t in {1,...,28}{
  \draw (\t,0.12) -- (\t,-0.12);
  \node[below=2pt,font=\scriptsize] at (\t,-0.12) {\t};
}

\foreach \a in {2,5,17}{
  \fill[red] (\a,0) circle (3pt);
}

\foreach \p in {1,4,9}{
  \draw[orange,fill=orange] (\p,0.18) circle (2.4pt);
}

\foreach \j/\lbl/\y/\yhat in {1/True Positive/1/1, 2/False positive/0/1, 3/False negative/1/0, 4/True negative/0/0}{
  \pgfmathsetmacro{\xstart}{7*(\j-1)+1}
  \pgfmathsetmacro{\xend}{7*(\j)}

  \draw[decorate,decoration={brace,mirror,amplitude=5pt}]
        (\xstart,-0.5) -- (\xend,-0.5);

  \node[below=17pt,font=\footnotesize]
        at ({(\xstart+\xend)/2},-0.5) {$W_{\j}$};

  \node[below=29pt,font=\scriptsize,align=center]
        at ({(\xstart+\xend)/2},-0.5)
        {$R_{\j}=\y,\;\hat R_{\j}=\yhat$\\(\lbl)};
}

\node[font=\scriptsize] at (14.5,-3.0) {%
  \tikz{\fill[red] (0,0.55ex) circle (3pt);}%
  \; Observed accident ($y_t = 1$)%
  \hspace{1.8cm}%
  \tikz{\draw[orange,fill=orange] (0,0.55ex) circle (2.4pt);}%
  \; Predicted accident ($\hat y_t = 1$)%
};

\end{tikzpicture}}

\caption{Weekly aggregation ($H = 7$) illustrating period-level risk.}
\label{fig:period-risk-4cases}
\end{figure*}
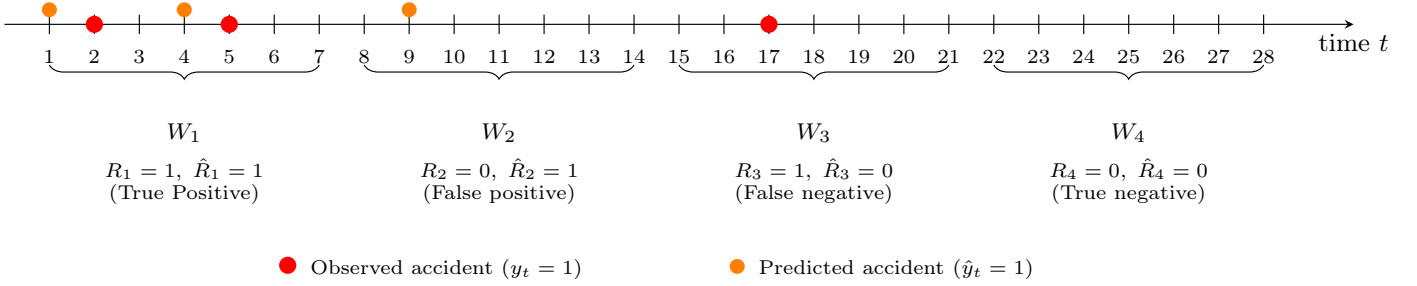

To implement this approach, we divide the observation horizon into consecutive periods of length $H$. For the $j$-th period, we define the index set $$W_j = \{(j-1)\cdot H + 1,\dots,j \cdot H\},$$ where $j \in \{1,\dots,P\} \text{ avec } P =  \lfloor \frac{T}{H} \rfloor.$
A binary variable $R_j$ is then introduced, which takes the value 1 if at least one accident occurs within the period, and 0 otherwise. The associated predictions, denoted $\hat{R}_j$, are defined analogously:
$$
  \hat{R}_j = \max_{t \in W_j}\,\hat{y}_t 
  \quad\text{and}\quad 
  R_j = \max_{t \in W_j}\,y_t.
$$ In the case where $H=7$, each period spans exactly one week, enabling analysis at a weekly scale. Figure \ref{fig:period-risk-4cases} shows this weekly segmentation.

To evaluate model performance under class imbalance, we compute several metrics that go beyond simple overall accuracy:

\emph{Recall (RE).} 
\[
  \text{RE} = \frac{\sum_{j=1}^{P} \mathbf{1}_{\{R_j = 1 \,\wedge\, \hat{R}_j = 1\}}}
                   {\sum_{j=1}^{P} \mathbf{1}_{\{R_j = 1\}}}.
\]
This metric quantifies the proportion of truly risky periods that are correctly detected, i.e., the model's ability to avoid missing actual accidents.

\emph{Precision (PR).} 
\[
  \text{PR} = \frac{\sum_{j=1}^{P} \mathbf{1}_{\{R_j = 1 \,\wedge\, \hat{R}_j = 1\}}}
                   {\sum_{j=1}^{P} \mathbf{1}_{\{\hat{R}_j = 1\}}}.
\]
This measures the proportion of periods predicted as risky that actually contained an accident, thus reflecting the reliability of alerts.

\emph{F1-score (F1).}
\[
  \text{F1} = 2 \,\cdot\, \frac{\text{PR} \cdot \text{RE}}{\text{PR} + \text{RE}}.
\]
The F1-score is the harmonic mean of precision and recall, emphasizing the balance between false alarms and missed accidents.

\emph{Specificity (SP).}
\[
  \text{SP} = \frac{\sum_{j=1}^{P} \mathbf{1}_{\{R_j = 0 \,\wedge\, \hat{R}_j = 0\}}}
                   {\sum_{j=1}^{P} \mathbf{1}_{\{R_j = 0\}}}.
\]
This metric captures the proportion of safe periods correctly classified as non-risky by the model.

\emph{Balanced Accuracy (BA).}
\[
  \text{BA} = \frac{\text{RA} + \text{SP}}{2}.
\]
Balanced accuracy is the average of recall and specificity, providing a global performance score particularly relevant under class imbalance conditions.

\section{Data description and preprocessing}

\subsection{Data description}

The dataset considered in this paper was collected from a company specializing in industrial waste management and covers the period from January 2019 to October 2022. Over this period, 2,108 safety inspections were conducted across 31 departments, and 479 accidents were recorded. During each visit, a feedback form was completed to document one or more hazardous situations, the actions required to remedy them, and any noteworthy best practices. In the dataset, The workers involved in recorded accidents are classified according to their contract type: they may be employees directly affiliated with the company, external personnel (with or without a specific contract), or temporary workers. These classifications have been consolidated into two main categories: internal and temporary worker (ITW), consisting of individuals directly affiliated with the company as well as temporary workers and external workers (ExW) consisting of personnel employed externally.

These feedback forms are filled out separately depending on whether they concern ITW or ExW. For ITW, data are collected at the departmental level. In contrast, ExW data are collected for the entire site rather than by department. In addition to analyzing the two principal categories (ITW and ExW), the present study also focuses on the department with the highest recorded accident rate, referred to here as Departement 1 (d1). Thus, ITW-d1 designates internal and temporary workers who experienced accidents in Departement 1.

\begin{table}[ht]
    \centering
    \setlength{\tabcolsep}{3pt}
    \begin{tabular}{lccc}
        \toprule
        \textbf{Statistics} & \textbf{ITW} & \textbf{ITW-d1} & \textbf{ExW} \\
        \midrule
        \textbf{Number of accidents} & \textbf{325} & \textbf{66} & \textbf{154} \\
        Number of safety inspections & 1770 & 597 & 336 \\
        Number of hazardous situations & 1392 & 468 & 302 \\
        Number of improvement actions & 1832 & 506 & 330 \\
        Number of best practices & 1319 & 447 & 265 \\
        \bottomrule
    \end{tabular}
    \caption{Accident, safety inspection, and best practice statistics for ITW, ITW-d1 and ExW}
    \label{table:1}
\end{table}

\begin{figure*}[t]
    \centering
    \includegraphics[scale=0.7]{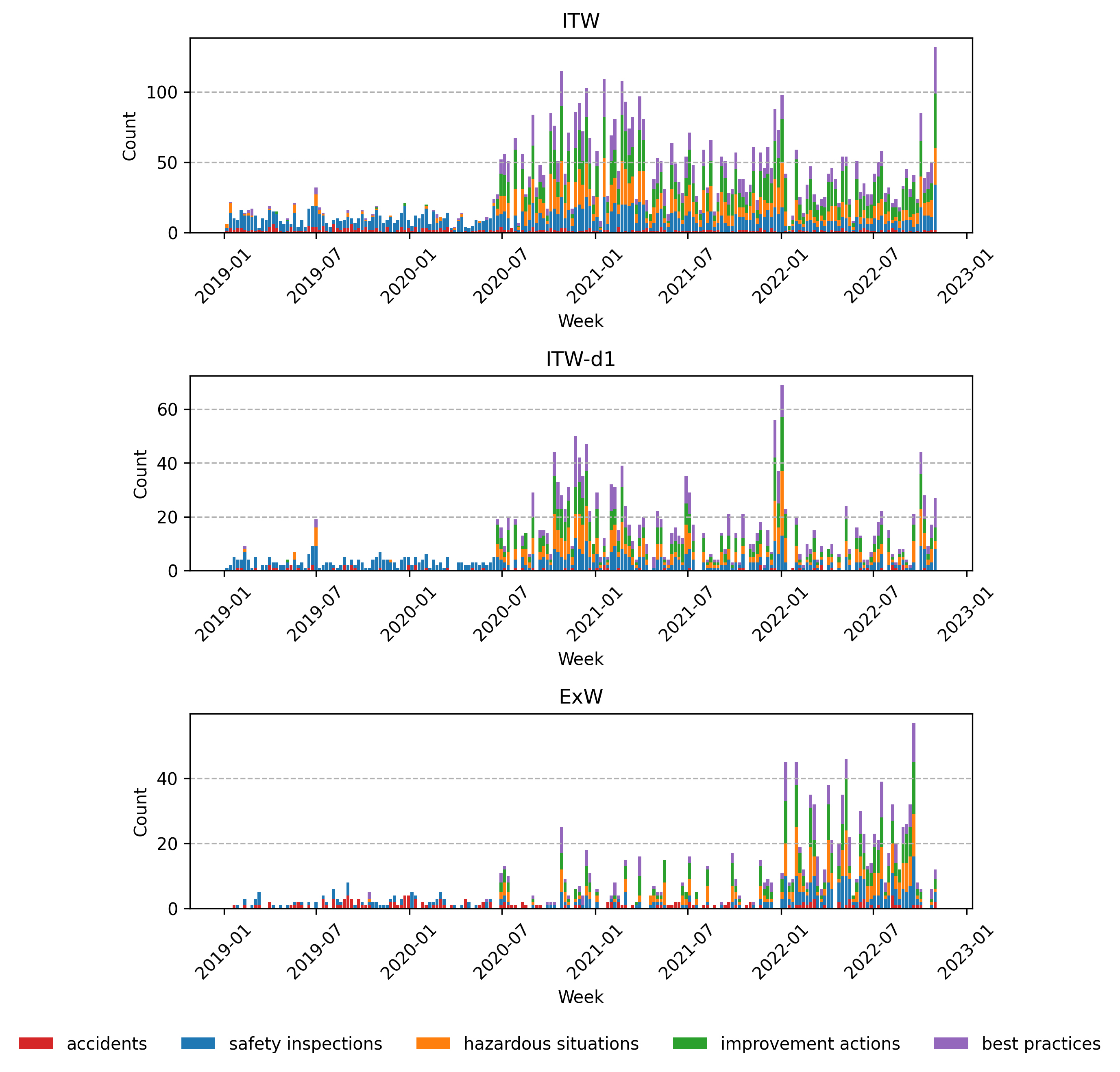}
    \caption{Number of weekly accidents, safety inspections, and best practices for ITW, ITW-d1, and ExW.}
    \label{fig:Figure_2}
\end{figure*}

Throughout the period of study, regular safety inspections were conducted, as shown in Figure~\ref{fig:Figure_2}. However, few observations were reported in 2019. From July 2020 onward, there was a marked increase in reported information for ITW, whereas ExW reports remained sparse until early 2021, when data collection intensified again. This lag likely reflects the impact of the COVID-19 pandemic, during which fewer external personnel were on-site, reducing the number of field observations recorded for external workers.

Overall, the total number of identified actions exceeds that of both hazardous situations and best practices across the observation period (see Table~\ref{table:1}). This trend suggests that whenever a hazardous situation is detected, several remedial actions are usually proposed, indicating a proactive approach to risk management. It also illustrates that, although highlighting best practices is important, the primary emphasis has been on defining and deploying targeted measures to mitigate accident risks.

\section{Data preparation and exploratory analysis}

\subsection{Variable overview and feature engineering}

 Safety inspection reports are predominantly textual, comprising descriptions of hazardous situations, recommended corrective actions, and best practices. To incorporate this information into our predictive models, it was converted into structured numerical indicators that may serve as early warning signals for potential accidents. Table~\ref{table:variables} provides a complete overview of the variables used in our study. The dataset includes variables that record daily events such as the number of hazardous situations reported per day, the median severity of hazardous situations per day and number of good practices per day. The binary outcome is derived from the number of accidents per day, and takes the value 1 if at least one accident occurred on a given day, and 0 otherwise. Table~\ref{table:variables} provides a complete overview of the variables used in our study.

\begin{table*}[t]
    \centering
    \renewcommand{\arraystretch}{1.2}
    \setlength{\tabcolsep}{3pt}
    \resizebox{0.85\textwidth}{!}{%
    \begin{tabular}{>{\ttfamily}l l >{\centering\arraybackslash}p{11cm}}
        \toprule
        \textbf{Variable} & \textbf{Type} & \textbf{Description} \\
        \midrule
        num\_accidents & discrete & Number of accidents per day \\
        num\_hazardous\_situations & discrete & Number of hazardous situations reported per day \\
        num\_improvement\_actions & discrete & Number of improvement actions per day \\
        num\_best\_practices & discrete & Number of good practices per day \\
        num\_safety\_inspections & discrete & Number of safety inspections per day \\
        severity\_median & categorical & Median severity of hazardous situations \\
        cleanliness\_median & categorical & Median cleanliness level of inspected areas \\
        days\_off\_median & discrete & Maximum number of days lost to work stoppages per day \\
        improvement\_progress\_median & categorical & Median initial progress rate of improvement actions \\
        month & categorical & Month of the inspection \\
        day\_of\_week & categorical & Day of the week \\
        quarter & categorical & Quarter of the year \\
        semester & categorical & Semester \\
        weekend\_day & categorical & Weekend indicator (saturday or sunday) \\
        holiday & categorical & Indicator for the summer break and public holidays\\
        \bottomrule
    \end{tabular}%
    }
    \caption{Description of the variables present in the dataset.}
    \label{table:variables}
\end{table*}

\subsection{Visual assessment of stationarity}

\begin{figure*}[t]
    \centering
    \includegraphics[scale=0.7]{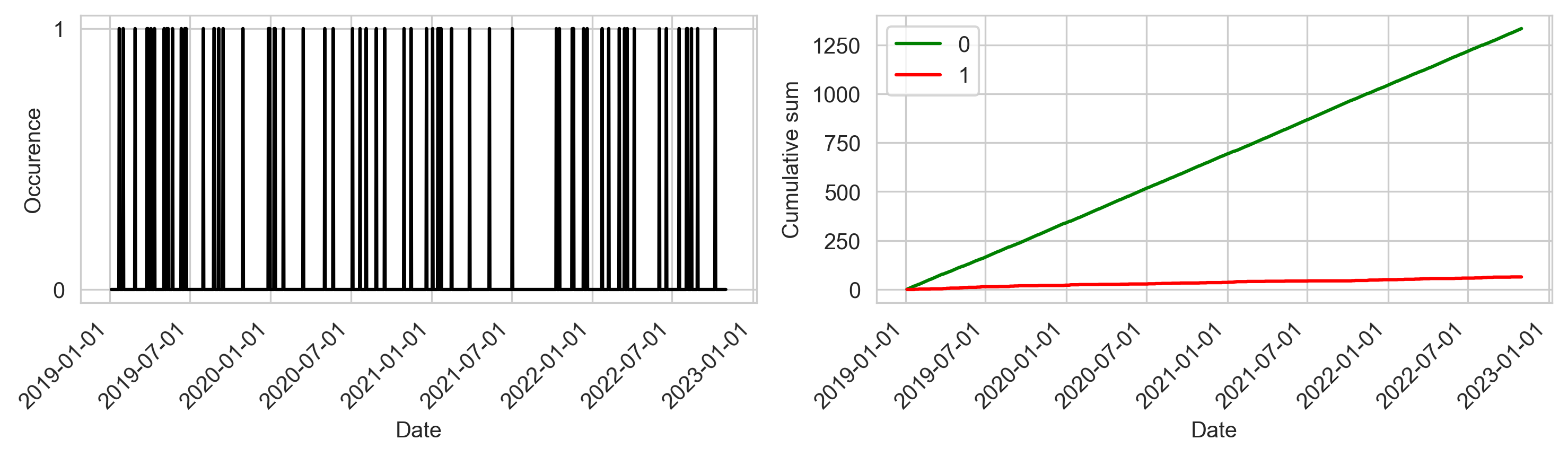}
    \caption{Binary time series (left) and rate evolution graph (right) of ITW-d1 series.}
    \label{fig:Figure_3}
\end{figure*}

Various visualization techniques exist for binary time series \citep[see][]{weiss2008visual}. One of the most effective is the \textit{rate evolution graph} \citep{ribler1997visualizing}, shown on the right-hand side of Figure~\ref{fig:Figure_3}.

In the binary case, this graph is constructed as follows: for $i \in \{0,1\}$, define the cumulative sums
$$
S_t^{(i)} \;=\; \sum_{s=1}^t \mathbf{1}_{\bigl\{X_s = i\bigr\}},$$
where $X_s$ denotes the binary outcome at time \(s\). 

The slope of the curve \(S_t^{(i)}\) provides an estimate of the marginal probability associated with outcome $i$. If the slopes of the two curves (for $i=0$ and $i=1$) remain fairly constant and linear, it suggests temporal stability in the marginal probabilities.

The cumulative curves for the ITW-d1 series (Figure~\ref{fig:Figure_3}) show nearly linear trends, suggesting that the series is stationary. The frequency of binary outcomes appears stable over time, a pattern that is also observed in the other series (see~\ref{Plot_series}).

\subsection{Autocorrelation and calendar effects}

\begin{figure}[t]
\centering
\includegraphics[scale=0.46]{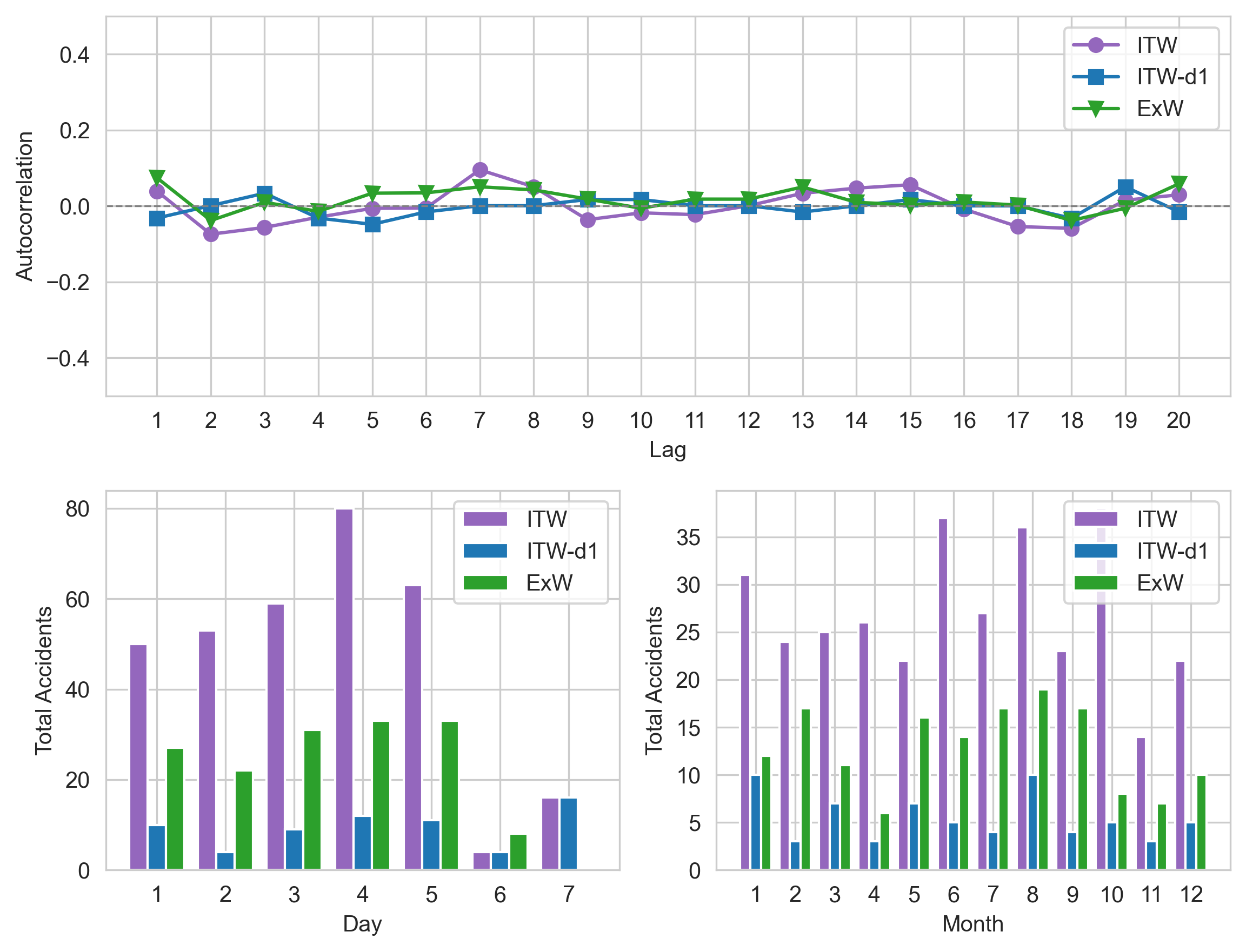}
"\caption{Autocorrelation and calendar-related effects in occupational accidents: autocorrelation (top), distribution by day of the week (bottom left), and by month (bottom right).}
\label{fig:Figure_4}
\end{figure}

Figure~\ref{fig:Figure_4} presents, in the top panel, the temporal dependence of the binary series as measured by Cohen's \textit{Kappa} statistic \citep{WeissGoeb2008,Weiss2009}. The autocorrelation remains weak at all lags, suggesting no strong memory effect from one day to the next.

The bottom figure illustrate how accidents are distributed over weekly and monthly time scales. Accidents tend to occur more frequently during midweek across all groups and decline noticeably over the weekend. The ITW, ITW-d1, and ExW series each exhibit specific patterns: ITW shows a slight peak in accident frequency during summer months; ITW-d1 displays a relatively uniform distribution throughout the year, with a marked drop on Saturdays and Sundays; and ExW stands in between, with occasional midweek peaks and a moderate increase during summer. Overall, while no strong seasonal effect emerges, these fluctuations suggest that weekly work rhythms and operational contexts specific to each group may influence the timing of accident occurrences.

\section{Training setup and evaluation}

The training period spans from 2019 to 2022, while data from 2022 onward were used for testing. This temporal split ensures that all evaluations are performed in a true out-of-sample setting, simulating real-world forecasting conditions. Table~\ref{tab:stats_data} summarizes the number of samples and the class distribution for each series across training and test sets. All three datasets (ITW, ITW-d1, and ExW) exhibit a strong class imbalance, with most days involving no accident. For instance, in the ITW-d1 test set, only 14 out of 289 days recorded an accident, i.e., less than 5\%. This imbalance highlights the importance of using evaluation metrics robust to rare events and using class-weighted loss functions to improve sensitivity to the minority class \citep{He2009Imbalanced}.
Model training and hyperparameter calibration, including the classification threshold, follow the time series cross-validation procedure described in Algorithm~\ref{algo:TSCV}. The initial training window covers 60\% of the training data, and the prediction horizon is set to 7 days. The model is retrained every 7 days to ensure up-to-date forecasting.For reproducibility, \ref{app:hp_grids} reports the hyperparameter search spaces and the selected operating points for all models considered in this study for the ITW, ITW-d1, and ExW.

\begin{table}[t]
\centering
\setlength{\tabcolsep}{6pt}
\begin{tabular}{l c cc cc}
    \toprule
    \textbf{Series} & \textbf{$T$}
    & \multicolumn{2}{c}{\textbf{Train}} 
    & \multicolumn{2}{c}{\textbf{Test}} \\
    & & \textbf{0} & \textbf{1} & \textbf{0} & \textbf{1} \\
    \midrule
    ITW     & 1397 & 863 & 231 & 258 & 45 \\
    ITW-d1  & 1397 & 1044 & 50 & 289 & 14 \\
    EXW     & 1397 & 990 & 104 & 268 & 35 \\
    \bottomrule
\end{tabular}
\caption{Train/test sample sizes and binary class distribution}
\label{tab:stats_data}
\end{table}

Hyperparameter tuning and model robustness assessment were performed using a sliding-window time series cross-validation (TSCV) approach \citep{Tashman2000, bergmeir2012crossval, hyndman2018forecasting}. This procedure consists in chronologically splitting the data into successive training and validation windows, thereby replicating realistic forecasting conditions. It offers more reliable out-of-sample performance estimates and is particularly well-suited for time series data.
In this study, TSCV is used at two key stages: first, for hyperparameter tuning based solely on the training data, as described in Algorithm~\ref{algo:TSCV}, and second, for the final model evaluation using an independent test set. During final evaluation, the initial training window includes the full training set, and previously tuned hyperparameters are used.

\begin{algorithm}[htbp]
\KwIn{%
  \begin{itemize}[noitemsep,nolistsep]
    \item Training set $\mathcal{D}_{\text{train}} = \{(x_i, y_i)\}_{i=1}^{N_{\text{train}}}$
    \item Initial window length $m$
    \item Step size $h$
    \item Grid of hyperparameter configurations $\Theta$
    \item Evaluation function $\operatorname{Metric}(\cdot)$
  \end{itemize}
}
\KwOut{\,Optimal hyperparameters $\theta^{\star}$}
Compute the number of validation folds: $K \leftarrow \left\lfloor \dfrac{N_{\text{train}} - m}{h} \right\rfloor$\;

\ForEach{$\theta \in \Theta$}{
  $\mathcal{P} \leftarrow \varnothing$\\
  \For{$k \gets 0$ \KwTo $K-1$}{
      $\mathcal{D}_{\text{train}}^{(k)} \leftarrow \{(x_i,y_i)\}_{i=1}^{m+kh}$\;
      $\mathcal{D}_{\text{val}}^{(k)} \leftarrow \{(x_i,y_i)\}_{i=m+kh+1}^{m+(k+1)h}$\;

      Train the model with $\theta$ on $\mathcal{D}_{\text{train}}^{(k)}$\;
      Predict $\hat y_i$ for each $(x_i,y_i)\in\mathcal{D}_{\text{val}}^{(k)}$\;

      $\mathcal{P} \leftarrow \mathcal{P} \cup \{(y_i,\hat y_i)\}_{(y_i,\hat y_i)\in\mathcal{D}_{\text{val}}^{(k)}}$\;
  }
  $\text{score}(\theta) \leftarrow \operatorname{Metric}(\mathcal{P})$\;
}
$\theta^{\star} \leftarrow \arg\max_{\theta\in\Theta} \text{score}(\theta)$\;
\caption{Sliding‑window cross‑validation.}
\label{algo:TSCV}
\end{algorithm}

\vspace{1em}
Algorithm~\ref{algo:TSCV} outlines the step-by-step validation process: the training window is gradually extended over time, while predictions are validated on the next window, providing a realistic short-term forecast assessment. In our study, the step size $H$ which defines the validation window length is set to 7 days to yield weekly forecasts.

\section{Results} 
\label{sec:results}

\subsection{Operational baselines}\label{sec:op-baselines}
To anchor the evaluation in practical routines, we benchmarked the learning models against two simple, operations-focused, training-free baselines reflecting rules commonly used by safety teams.

\subsubsection{Rolling accident frequency over $W$ days}

Following standard safety monitoring practices, the \textit{rolling frequency baseline} approximates the short-term accident rate over the last \(W\) days. This design is conceptually aligned with the \textit{total recordable incident rate (TRIR)}, a well-established safety indicator that aggregates recent event counts over fixed time windows to track variations in safety performance \citep{hse2001measuring}. Such trailing indicators are consistent with operational safety management approaches, where repeated minor incidents or a rise in local accident frequency are regarded as early warning signals of deteriorating safety conditions \citep{goh2012organizational,reason1997managing}. 

A trailing accident rate is computed
\[
r_t=\frac{1}{W}\sum_{i=0}^{W-1} y_{t-i},
\]
and we use it as a constant risk score for the whole upcoming week:
\[
\hat{p}_{t+h}=r_t,\qquad h=1,\dots,7.
\]

To convert this score into a period-level alert, we flag the upcoming week as \emph{at risk} whenever \(r_t \ge \tau\).
The threshold \(\tau\) and the window length \(W\) (chosen from \(\{7,14,28\}\)) are selected on the training split by maximizing period-level balanced accuracy, and then held fixed for evaluation.This procedure yields \(W{=}14\) and thresholds \(\tau_{\text{ITW}}{=}0.4\), \(\tau_{\text{ITW-d1}}{=}0.4\), and \(\tau_{\text{EXW}}{=}0.5\).

\subsubsection{Naive}
This baseline copies last week’s pattern into the next week:
\[
\hat{y}_{t+h}=y_{t+h-7}, \qquad h=1,\dots,7.
\]
It reflects the intuition that short-term risk often repeats on the same weekday.

\subsection{Model performance assessment}
Table~\ref{tab:model_evaluation} compares all methods on the three series using Recall (RE), Precision (PR), F1, and balanced accuracy (BA). Across datasets, learned models consistently outperform the operational baselines, confirming the usefulness of the proposed framework. On ITW and ITW-d1, the LSTM (MIMO) achieves the highest BA, indicating superior ability to detect risky weeks while limiting false alarms. On ExW, a simple Decision Tree attains the best BA, suggesting that a low-capacity, rule-like boundary fits that series’ patterns particularly well. Tree/boosting models and the MLP are competitive but generally below the best deep sequence model on ITW and ITW-d1. The rolling-frequency baseline ({\emph{Freq.\ rolling}) remains informative yet is surpassed by most learned models, which better exploit temporal dependencies and calendar effects. Across the three series, LSTM--Seq2seq, TCN, and TFT are consistently competitive. 
On ITW and ITW-d1, they generally rank below the top LSTM--MIMO yet clearly above the operational baselines, and they are broadly comparable to the stronger tree/boosting and MLP models. 
On ExW, none of the three sequence learners takes the lead; the best score is obtained by a Decision Tree, while the sequence models remain competitive. 

\begin{table*}[t]
    \centering
    \scriptsize
    \renewcommand{\arraystretch}{1.3}
    \setlength{\tabcolsep}{4pt}
    \begin{tabular}{ll
                    cccc
                    cccc
                    cccc}
        \toprule
        \multirow{2}{*}{\textbf{Model family}} &
        \multirow{2}{*}{\textbf{Model}} &
            \multicolumn{4}{c}{\textbf{ITW}} &
            \multicolumn{4}{c}{\textbf{ITW-d1}} &
            \multicolumn{4}{c}{\textbf{ExW}} \\
        \cmidrule(lr){3-6}\cmidrule(lr){7-10}\cmidrule(lr){11-14}
            & & \textbf{RE} & \textbf{PR} & \textbf{F1} & \textbf{BA} &
                \textbf{RE} & \textbf{PR} & \textbf{F1} & \textbf{BA} &
                \textbf{RE} & \textbf{PR} & \textbf{F1} & \textbf{BA} \\
        \midrule

        \multirow{2}{*}{Operational baselines}
        & Naive (last week)
        & 0.66 & 0.68 & 0.67 & \cellcolor{CAD7E5}0.37
        & 0.15 & 0.15 & 0.15 & \cellcolor{CAD7E5}0.39
        & 0.56 & 0.58 & 0.57 & \cellcolor{CAD7E5}0.50 \\
        & Freq.\ rolling (W)
        & 0.69 & 0.79 & 0.73 & \cellcolor{CAD7E5}0.57
        & 0.08 & 0.33 & 0.12 & \cellcolor{CAD7E5}0.50
        & 0.52 & 0.65 & 0.58 & \cellcolor{CAD7E5}0.56 \\
        \midrule

        \multirow{7}{*}{Machine learning}
        & Logistic regression
        & 0.37 & 1.00 & 0.54 & \cellcolor{CAD7E5}0.69
        & 0.46 & 0.75 & 0.57 & \cellcolor{CAD7E5}0.70
        & 0.78 & 0.67 & 0.72 & \cellcolor{CAD7E5}0.67 \\
        & Linear discriminant analysis
        & 0.59 & 0.79 & 0.68 & \cellcolor{CAD7E5}0.57
        & 0.50 & 0.29 & 0.36 & \cellcolor{CAD7E5}0.51
        & 0.88 & 0.66 & 0.76 & \cellcolor{CAD7E5}0.63 \\
        & Decision tree
        & 0.81 & 0.87 & 0.84 & \cellcolor{CAD7E5}0.72
        & 0.46 & 0.66 & 0.54 & \cellcolor{CAD7E5}0.68
        & \textbf{0.64} & \textbf{0.84} & \textbf{0.73} & \cellcolor{CAD7E5}\textbf{0.74} \\
        & Random forest
        & 0.62 & 0.82 & 0.71 & \cellcolor{CAD7E5}0.66
        & 0.92 & 0.38 & 0.54 & \cellcolor{CAD7E5}0.64
        & 0.88 & 0.65 & 0.74 & \cellcolor{CAD7E5}0.61 \\
        & Histogram boosting gradient
        & 0.93 & 0.81 & 0.87 & \cellcolor{CAD7E5}0.65
        & 0.38 & 0.71 & 0.50 & \cellcolor{CAD7E5}0.66
        & 0.28 & 0.87 & 0.42 & \cellcolor{CAD7E5}0.61 \\
        & XGBoost
        & 0.69 & 0.88 & 0.77 & \cellcolor{CAD7E5}0.71
        & 0.85 & 0.46 & 0.59 & \cellcolor{CAD7E5}0.71
        & 0.48 & 0.80 & 0.60 & \cellcolor{CAD7E5}0.67 \\
        & LightGBM
        & 0.78 & 0.86 & 0.81 & \cellcolor{CAD7E5}0.70
        & 0.92 & 0.43 & 0.58 & \cellcolor{CAD7E5}0.69
        & 0.76 & 0.65 & 0.70 & \cellcolor{CAD7E5}0.60 \\
        \midrule

        \multirow{5}{*}{Deep learning}
        & Multi-layer perceptron
        & 0.62 & 0.90 & 0.74 & \cellcolor{CAD7E5}0.72
        & 0.69 & 0.47 & 0.56 & \cellcolor{CAD7E5}0.68
        & 0.64 & 0.76 & 0.69 & \cellcolor{CAD7E5}0.68 \\
        & \textbf{LSTM -- MIMO}
        & \textbf{0.71} & \textbf{0.92} & \textbf{0.80} & \cellcolor{CAD7E5}\textbf{0.73}
        & \textbf{0.85} & \textbf{0.79} & \textbf{0.82} & \cellcolor{CAD7E5}\textbf{0.87}
        & 0.79 & 0.70 & 0.75 & \cellcolor{CAD7E5}0.67 \\
        & Seq2seq
        & 0.69 & 0.88 & 0.77 & \cellcolor{CAD7E5}0.70
        & 0.53 & 0.70 & 0.60 & \cellcolor{CAD7E5}0.72
        & 0.64 & 0.69 & 0.66 & \cellcolor{CAD7E5}0.62 \\
        & TCN
        & 0.65 & 0.91 & 0.76 & \cellcolor{CAD7E5}0.71
        & 0.92 & 0.50 & 0.51 & \cellcolor{CAD7E5}0.76
        & 0.52 & 0.76 & 0.62 & \cellcolor{CAD7E5}0.65 \\
        & TFT
        & 0.62 & 0.83 & 0.71 & \cellcolor{CAD7E5}0.63
        & 0.91 & 0.39 & 0.55 & \cellcolor{CAD7E5}0.68
        & 0.13 & 1.00 & 0.23 & \cellcolor{CAD7E5}0.56 \\
        \bottomrule
    \end{tabular}
    \caption{Performance comparison on the ITW, ITW-d1 and ExW series on the test set.}
    \label{tab:model_evaluation}
\end{table*}

\begin{figure*}[t]
\centering
\includegraphics[scale=0.5]{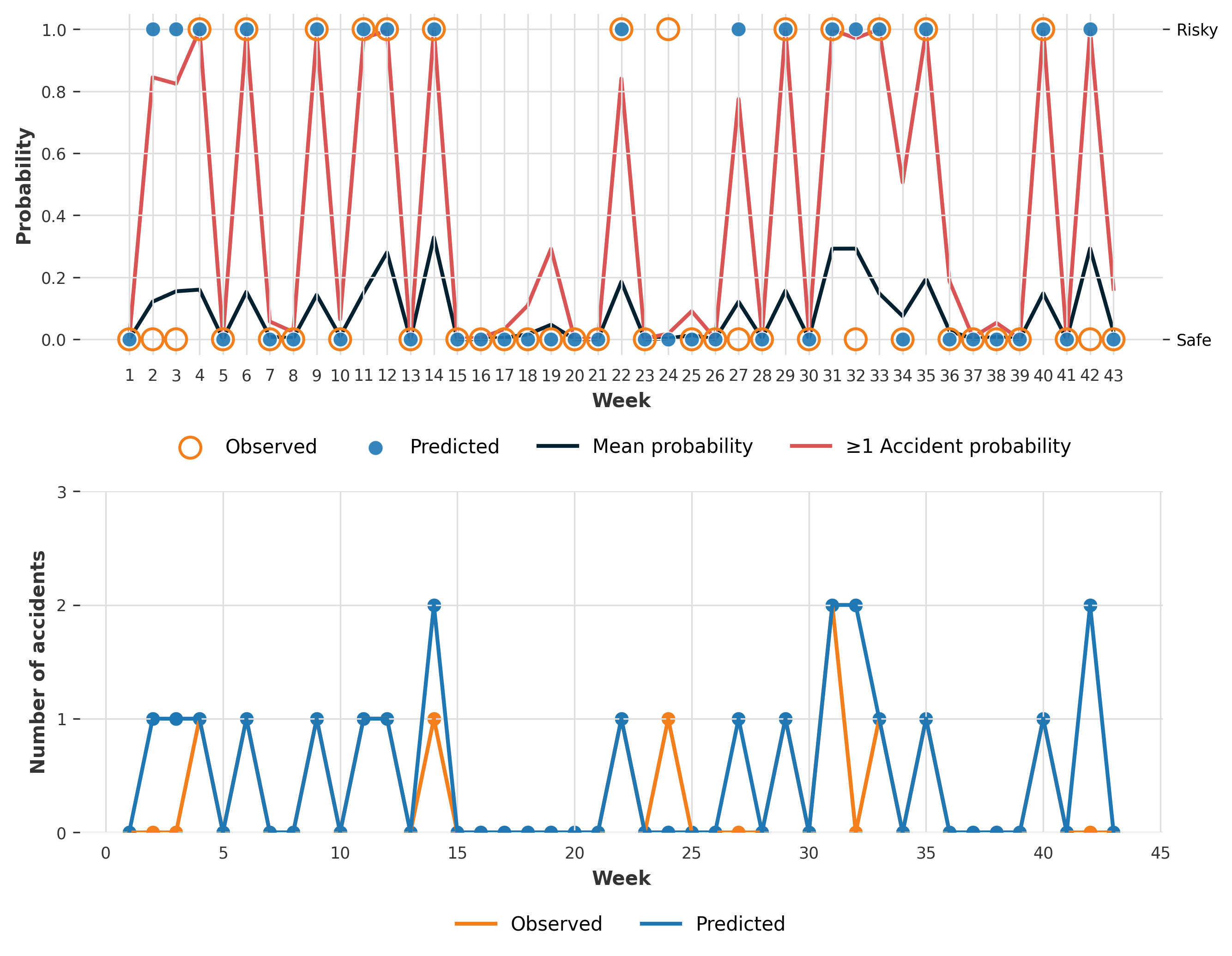}
\caption{Weekly accident risk prediction (top) and comparison with observed accident counts (bottom) on the ITW-d1 series with LSTM-MIMO model.}
\label{fig:Figure_5}
\end{figure*}

Figure~\ref{fig:Figure_5} illustrates these findings at the weekly level. The top panel shows a close alignment between observed and predicted risky weeks; remaining false positives occur when the model assigns high probabilities close to the decision threshold, which can still be valuable as preventive alerts. The bottom panel compares weekly accident counts and shows that the model tracks week-to-week variations reasonably well, with mild overestimation in a few weeks that remains acceptable in a prevention setting.

\begin{figure}[t]
\centering
\includegraphics[scale=0.5]{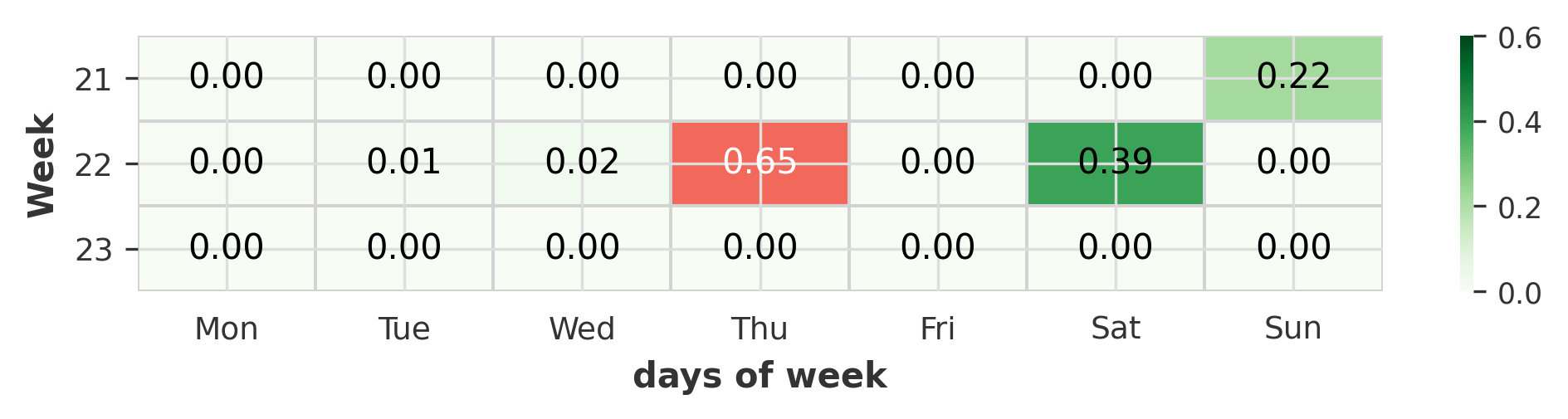}
\caption{Daily accident probabilities for Weeks 21,22 and 23.  
           Red cells indicate probabilities $>0.6$}
\label{tab:daily_probs_3weeks}
\end{figure}

Table \ref{tab:daily_probs_3weeks} provides a detailed focus on three consecutive weeks from the ITW-d1 series (results for the other series are available in \ref{forecast_series}). These weeks are classified respectively as safe, risky, and safe. The risky week contain one day with high predicted probability, supporting their classification. In contrast, the safe week is clearly identified, showing no alarming signals. This detailed view confirms the  ability of our framework to  dynamically capture short-term shifts in risk and to adapt to evolving safety conditions.

\subsection{Horizon sensitivity and calendar effects}
\label{sec:horizon-calendar}

Here, We assess how the decision period length or horizon (\(h\in\{3,5,7\}\)) impacts period-level performance and how known future calendar covariates \(s_{t+h}\) impact the performance of our models. 

\subsubsection{Horizon sensitivity}

Figure~\ref{fig:horizon_ba_all} shows that balanced accuracy improves with longer horizons, peaking at the weekly horizon \(\,h=7\,\) across all series. This pattern can be explained by two complementary effects: aggregating decisions over seven days smooths the noise inherent to rare events and yields more stable period-level signals, and several operational drivers (for example, inspection cadence and work scheduling) naturally follow weekly rhythms. The largest improvement is observed for ITW-d1, where events are concentrated in a single department and weekly aggregation therefore provides particularly informative signals.

From an operational perspective, a one-week look-ahead constitutes a natural planning unit: it affords sufficient lead time to allocate resources (e.g., targeted safety briefings or housekeeping actions), schedule interventions, and communicate priorities, while remaining specific enough to be actionable. Shorter horizons (\(h=3\) or \(h=5\)) can be more timely but tend to be noisier and less aligned with organisational processes, which explains their lower period-level balanced accuracy.

\begin{figure}[t]
\centering
\includegraphics[scale=0.46]{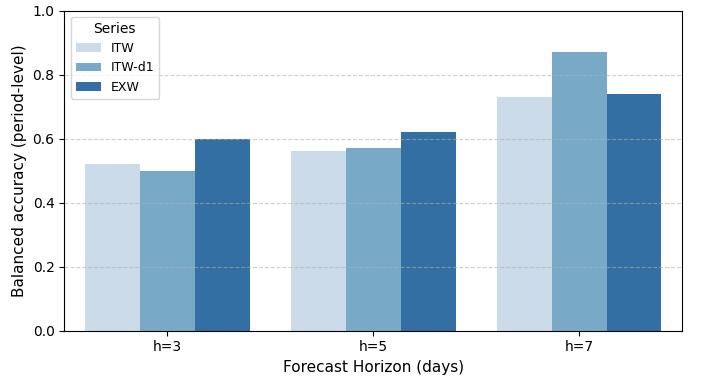}
"\caption{\textbf{Period-level Balanced accuracy by forecast Horizon (Best model per series).}}
\label{fig:horizon_ba_all}
\end{figure}

\subsubsection{Calendar ablation at weekly horizon.}

To isolate the marginal contribution of calendar features at \(h{=}7\), we rerun the identical backtesting protocol after \emph{removing} \(s_{t+h}\) from the future covariates, keeping all else fixed (same folds, same training data, same optimization settings, and the operating threshold \(\tau\) fixed to the value calibrated in the full model). Let \(\mathrm{BA}^{(\text{cal})}(D)\) denote the period-level balanced accuracy on dataset \(D\) \emph{with} calendar features, and \(\mathrm{BA}^{(\neg\text{cal})}(D)\) the BA \emph{without} them. We report the absolute gain
\[
\Delta\mathrm{BA}(D) \;=\; \mathrm{BA}^{(\text{cal})}(D) - \mathrm{BA}^{(\neg\text{cal})}(D).
\]

\begin{table}[t]
\centering
\small
\renewcommand{\arraystretch}{1.2}
\setlength{\tabcolsep}{3pt}
\begin{tabular}{llccc}
\toprule
\textbf{Series} & \textbf{Model} & \textbf{with calendar} & \textbf{no calendar} & \textbf{$\Delta$BA} \\
\midrule
ITW    & LSTM--MIMO     & 0.71 & 0.59 & $+$0.12 \\
ITW-d1 & LSTM--MIMO     & 0.87 & 0.61 & $+$0.26 \\
ExW    & Decision tree  & 0.74 & 0.64 & $+$0.10 \\
\bottomrule
\end{tabular}
\caption{Period-level balanced accuracy (BA) at \(h{=}7\) for the best model of each series (selected according to BA), with and without future calendar covariates.}
\label{tab:ba_h7_calendar}
\end{table}

Consistent, positive \(\Delta\mathrm{BA}\) across all series indicates that weekday/weekend and holiday structure is informative at the weekly scale (Table~\ref{tab:ba_h7_calendar}). The gain is lower for ITW ($+0.12$) and EXW ($+0.10$), suggesting that calendar signals are diluted when aggregating multiple departments with heterogeneous rhythms and that short-term history already captures weak weekly recurrences. In contrast, ITW-d1 shows a large improvement ($+0.26$), consistent with a more regular weekly pattern within a single department where calendar cues are highly predictive.

\section{Discussion}
\label{sec:discussion}

The proposed framework aims at supporting proactive accident prevention by providing weekly assessments of risk levels based on recurrent safety inspections. After analyzing feedbacks from safety inspections, the system estimates the probability of an accident for each day. These probabilities are then compared to a calibrated threshold: if the probability exceeds this threshold, the day is flagged as \emph{at risk}. A week is classified as \emph{risky} if at least one day exceeds the threshold. The framework is model-agnostic: preprocessing, feature engineering, validation protocol and scoring procedures are applied identically across model families, enabling fair comparisons and easy substitution of prediction models.
In practical terms, the output is designed as decision support rather than automatic enforcement: the threshold can be tuned to favor either recall (capture more risky weeks with more false alarms) or precision (fewer alerts with higher confidence), reflecting each site's prevention policy and resource constraints.
Alerts translate into simple playbooks for prevention teams (e.g., schedule a targeted walk-down or conduct a focused toolbox talk on the flagged day). This makes the signal immediately actionable for weekly planning.
Rather than aggregating all accidents into a single binary label, we concentrate on one specific accident category defined in Appendix~A and apply the same weekly pipeline to this category only. The model produces category-specific daily probabilities; a day is flagged as \emph{at risk} for that category when its probability exceeds the calibrated threshold, and a week is deemed \emph{risky} for that category if at least one day is flagged. This category focus yields status labels that directly reflect the expected occurrence of that accident type and enables more targeted prevention actions. The weekly roll-up integrates naturally with leading indicators already tracked by HSE (inspection counts, hazardous situations, improvement actions): combining a \emph{risky} week flag with weak-signal trends helps prioritize scarce resources and improves the timeliness of preventive actions.

This setup enables safety teams to anticipate high-risk periods. In practice, the model can be used at the beginning of each week to generate a brief report indicating whether the upcoming week is considered safe or risky, and which specific days require attention. For example, if Week 4 is flagged as risky due to to day 1 and 2, the safety officer may schedule targeted actions such as site visits, safety briefings, or specific inspections on those days. Thanks to its time-based design, the model can be regularly updated with newly reported field data. It can also be easily integrated into a prevention dashboard with automatic updates (e.g., every Monday) to support weekly planning. The more frequently and consistently field reports are submitted, the more robust the model becomes capturing subtle shifts in weak signals and risky behaviors over time. Despite these advantages, the approach also presents certain limitations. Aggregating data at the weekly level may sometimes under represent risks spread across several moderately risky days. For instance, three consecutive days with medium-level probabilities might not trigger any alert, whereas a single day with a high probability could result in the whole week being labeled as risky. This simplification should be kept in mind when interpreting results. Moreover, the poor performance of some traditional models may stem from poorly calibrated probability outputs. If the predicted scores do not accurately reflect reality, it becomes difficult to set effective thresholds to differentiate safe from risky periods. Improving probability calibration enhance the reliability of the alerts. Several directions can be explored to improve the system. Better-calibrated probabilistic models could be developed; decision thresholds could be tailored to specific departments; more advanced architectures, such as sequence-to-sequence models, could be considered; and global forecasting could help capture cross-site dependencies. Beyond accuracy, future work will report cost-sensitive analyses contrasting the expected cost of a missed risky week versus the operational cost of investigating a false alarm, which is central to prevention impact. Additionally, the textual content of hazardous-situation reports, currently summarized through quantitative indicators such as severity or frequency, could be leveraged more thoroughly. Semantic analysis of descriptions (e.g., using natural language processing techniques) could extract richer and potentially predictive signals while better capturing the specific context of reported events.

\section{Conclusion}
\label{sec:conclusion}
This study introduced a model-agnostic and operational framework for short-term forecasting of occupational accidents based on binary time series. Using proactive data from safety inspections, the model dynamically predicts daily accident probabilities and provides a weekly classification of risk. The proposed approach demonstrated its effectiveness in identifying both risky and safe periods, particularly through the use of an LSTM model, which outperformed classical machine learning methods in evaluated series.
From a prevention perspective, the weekly flag and day-level probabilities offer a simple, auditable signal that can be embedded in standard routines (weekly safety meetings, planning boards, shift handovers) to prioritize inspections and briefings precisely when they are most needed.
Because the pipeline is model-agnostic and thresholds are tunable, sites can adopt operating modes that reflect their risk appetite and seasonality, while keeping a consistent validation protocol across algorithms.

Due to its temporal structure and weekly aggregation, this framework is well suited to support real-time prevention strategies in industrial settings. It can be integrated into a safety dashboard and updated regularly to help decision makers plan targeted actions. The ability to anticipate high-risk weeks opens new avenues for more proactive, data-driven safety management.
Future deployments will include calibration monitoring and drift checks, horizon-specific summaries for short maintenance windows, and category-aware dashboards that link each risky week to concrete action checklists. We anticipate these enhancements will further translate predictive accuracy into measurable reductions in incident rates and improved timeliness of preventive actions.

\clearpage
\onecolumn

\appendix

\section{Accident categories and injury nature}
\label{categories}

Table~\ref{fig:accident_categories} present accident profiles for the ITW, ExW, and ITW-d1 groups. In both the ITW and ExW groups, accidents related to products, emissions, and waste and those involving work equipment are the most common. In contrast, the ITW-d1 group shows a different pattern: same-level falls and pedestrian movement dominate at 36.36\%, followed by work equipment at 24.24\% and products/emissions at 22.73\%. This distribution suggests that workers in the ITW-d1 group face the highest risk of movement hazards.Regarding the nature of injuries (Table~\ref{fig:injury_nature}), distinct patterns emerge across the groups. In ITW, musculoskeletal pain accounts for 24\% of accidents, a tendency also observed in ITW-d1 at 16.67\%. In contrast, ExW reports a higher frequency of wounds at 19.61\%. Chemical burns rank among the most common injuries across all groups 15.69\% for ITW, 15.03\% for ExW, and 12.12\% for ITW-d1. Additionally, a notable share of accidents resulted in no apparent injury (16.92\% for ITW, 22.22\% for ExW, and 24.24\% for ITW-d1).

\begin{table}[H]
  \centering
  \begin{subtable}[t]{0.48\textwidth}
    \centering
    \setlength{\tabcolsep}{3pt}
    \begin{tabular}{>{\raggedright\arraybackslash}p{3cm}ccc}
      \toprule
      \textbf{Accident categories} & \textbf{ITW (\%)} & \textbf{ExW (\%)} & \textbf{ITW-d1 (\%)} \\
      \midrule
      Related to products, emissions and waste & 28.62 & 28.57 & 22.73 \\
      Work equipment                            & 28.00 & 25.97 & 24.24 \\
      Same-level falls and pedestrian movement  & 26.46 & 17.53 & 36.36 \\
      Physical workload                         & 4.92  & 5.84  & 1.52  \\
      Thermal environments                      & 4.31  & 4.55  & 12.12 \\
      Internal vehicle/machine traffic          & 3.38  & 3.25  & 0.00  \\
      Collapses and falling objects             & 2.77  & 7.14  & 0.00  \\
      Mechanical handling                       & 0.62  & 0.65  & 0.00  \\
      Electricity                               & 0.31  & 1.95  & 0.00  \\
      Fire, explosion                           & 0.31  & 0.00  & 1.52  \\
      Noise                                     & 0.31  & 0.00  & 1.52  \\
      Fall from height                          & 0.00  & 1.30  & 0.00  \\
      Pressurized equipment (fluids, gas)       & 0.00  & 1.95  & 0.00  \\
      Psychosocial factors                      & 0.00  & 1.30  & 0.00  \\
      \bottomrule
    \end{tabular}
    \caption{Accident categories for the ITW, ExW, and ITW-d1 series.}
    \label{fig:accident_categories}
  \end{subtable}
  \hfill
  \begin{subtable}[t]{0.48\textwidth}
    \centering
    \setlength{\tabcolsep}{3pt}
    \begin{tabular}{>{\raggedright\arraybackslash}p{3cm}ccc}
      \toprule
      \textbf{Injury Nature} & \textbf{ITW (\%)} & \textbf{ExW (\%)} & \textbf{ITW-d1 (\%)} \\
      \midrule
      Musculoskeletal pain     & 24.00 &  8.50 & 16.67 \\
      No injury                & 16.92 & 22.22 & 24.24 \\
      Chemical burn            & 15.69 & 15.03 & 12.12 \\
      Wound                    & 14.77 & 19.61 & 13.64 \\
      Physical shock           &  6.15 &  4.58 &  7.58 \\
      Discomfort               &  5.23 &  5.23 &  1.52 \\
      Hematoma                 &  3.38 &  2.61 &  1.52 \\
      Thermal burn             &  3.08 &  5.88 & 10.61 \\
      Limb twist               &  2.15 &  1.96 &  3.03 \\
      Crushing injury          &  1.85 &  2.61 &  1.52 \\
      Discomfort/faintness     &  1.85 &  4.58 &  3.03 \\
      Fracture                 &  1.85 &  1.31 &  3.03 \\
      Irritation               &  1.85 &  3.27 &  1.52 \\
      Low back pain            &  0.92 &  1.31 &  0.00 \\
      Poisoning                &  0.31 &  0.65 &  0.00 \\
      Electric shock           &  0.00 &  0.65 &  0.00 \\
      \bottomrule
    \end{tabular}
    \caption{Injury nature for the ITW, ExW, and ITW-d1 series.}
    \label{fig:injury_nature}
  \end{subtable}
\end{table}

\section{Visualization of binary time series and rate evolution graph of ITW et ExW series}
\label{Plot_series}

\begin{figure}[H]
    \centering
    \includegraphics[scale=0.4]{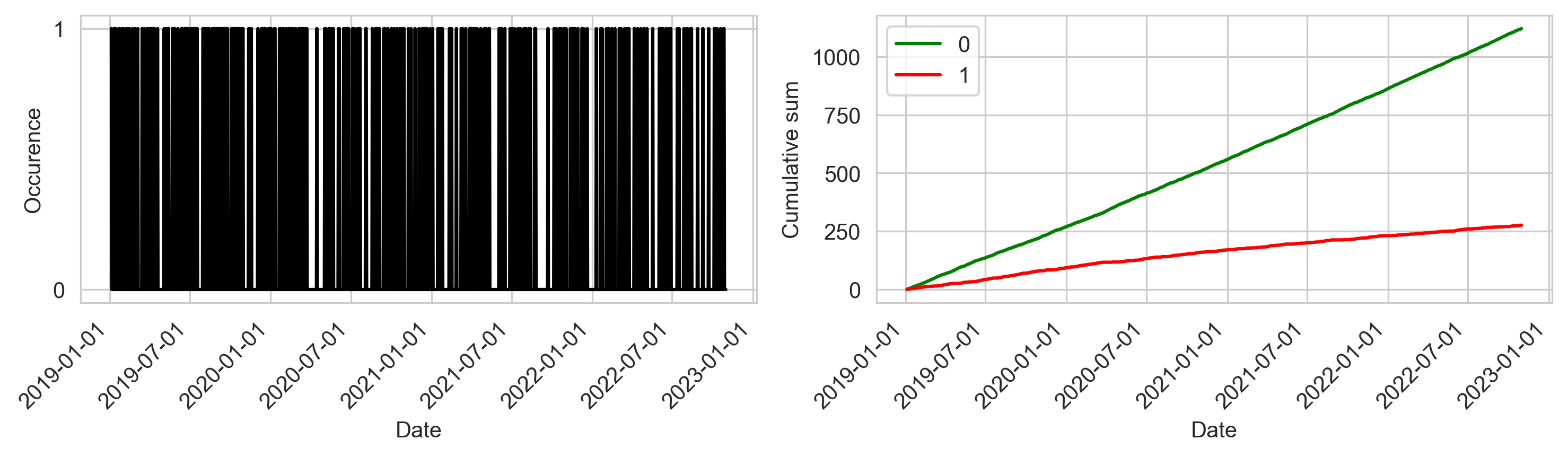}
    \caption{Binary time series (left) and rate evolution graph (right) of ITW-d1 series.}
    \label{fig:Figure_6}
\end{figure}

\begin{figure}[H]
    \centering
    \includegraphics[scale=0.4]{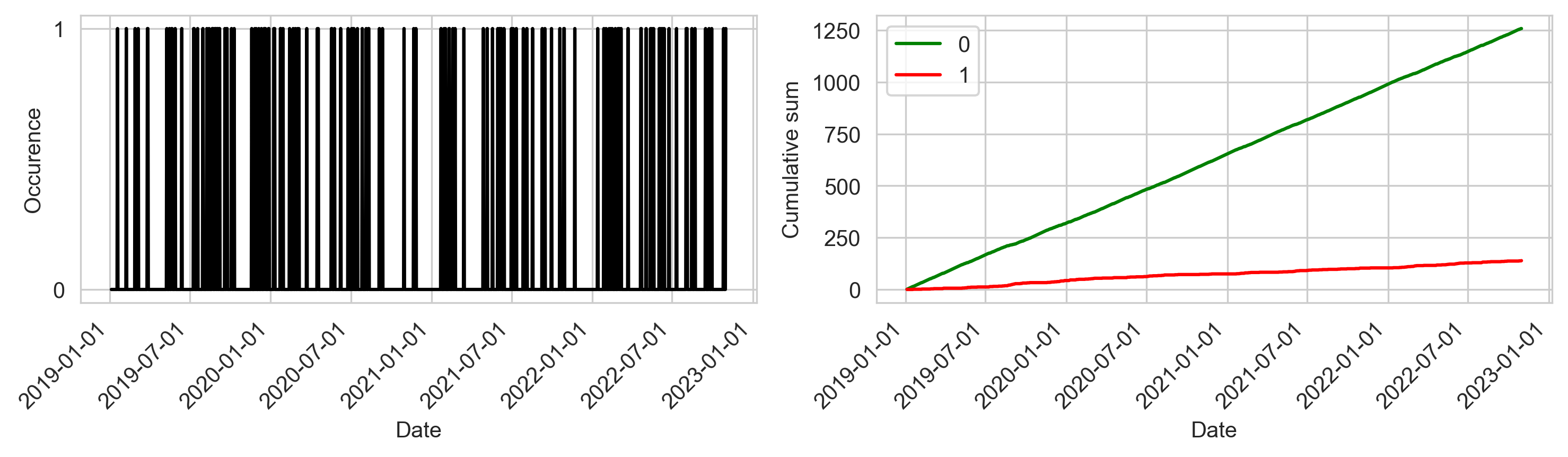}
    \caption{Binary time series (left) and rate evolution graph (right) of ExW series.}
    \label{fig:Figure_7}
\end{figure}

\section{Weekly accident risk prediction and comparison with observed accident counts on the ITW and ExW series}
\label{forecast_series}

\begin{figure}[H]
\centering
\includegraphics[scale=0.45]{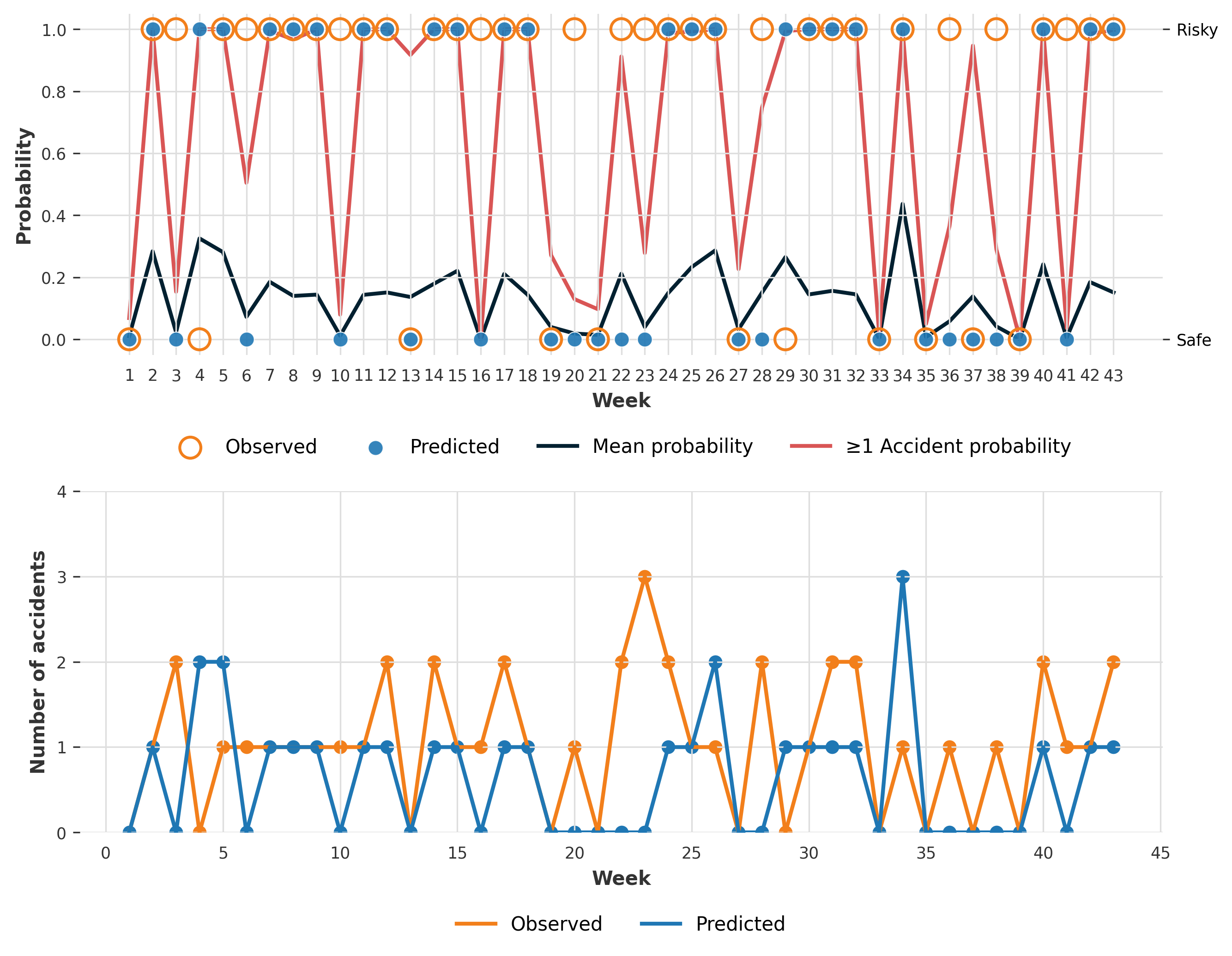}
\caption{Weekly accident risk prediction and comparison with observed accident counts on the ITW series with LSTM-MIMO model.}
\label{fig:Figure_8}
\end{figure}

\begin{figure}[H]
\centering
\includegraphics[scale=0.45]{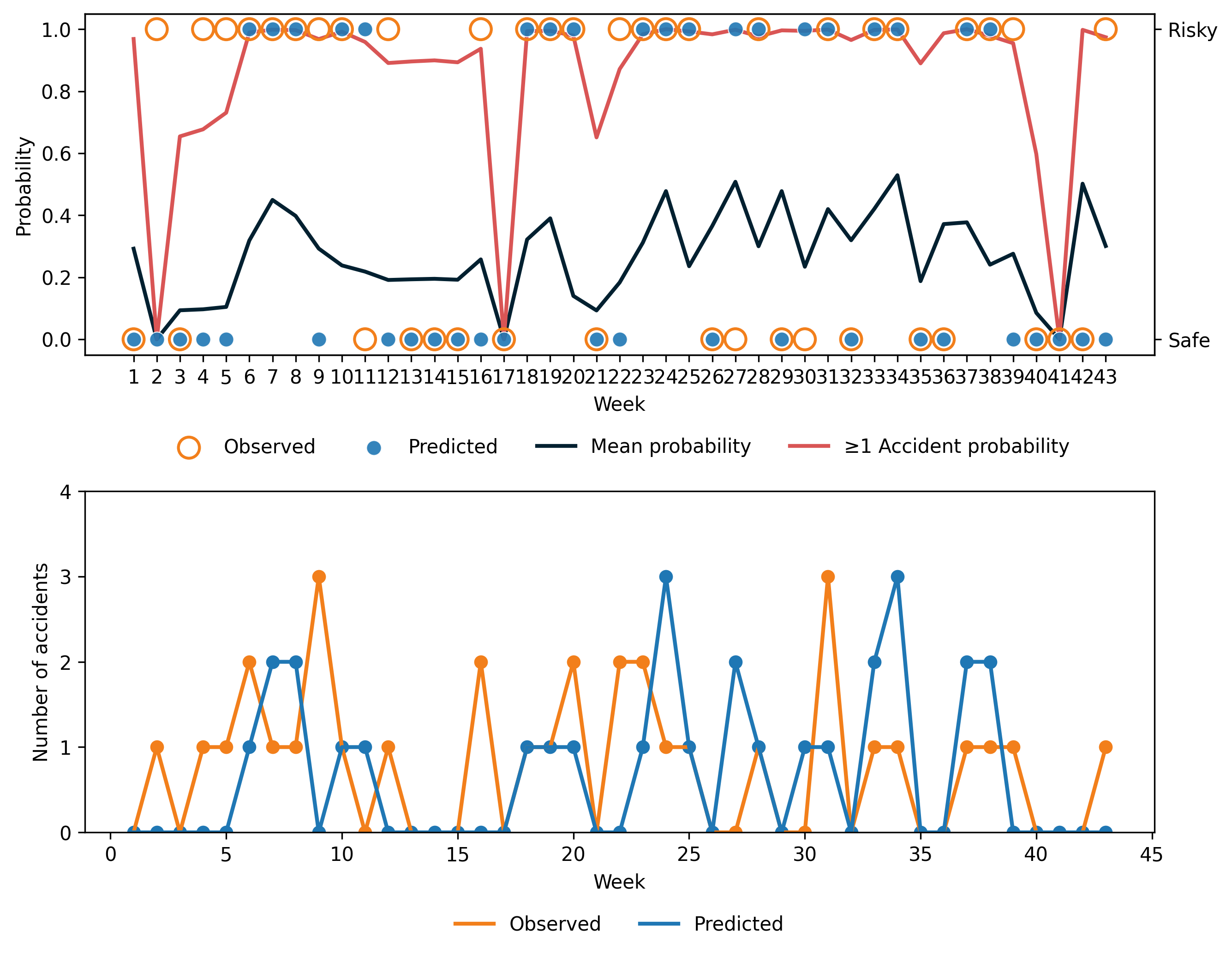}
\caption{Weekly accident risk prediction and comparison with observed accident counts on the ExW series with decision-tree model.}
\label{fig:Figure_9}
\end{figure}

\clearpage
\section{Hyperparameters grids and selected settings}
\label{app:hp_grids}

Table~\ref{tab:hp_results_ml} summarizes the grids and the settings selected by grid search for a weekly horizon \(H{=}7\). The lag depths (\(d_y\) and \(d_c\))  are tuned over \(\{7,14,28\}\). Implementations rely on \texttt{scikit-learn} and native libraries. For each dataset (ITW, ITW-d1, EXW), the table reports the best hyperparameters together with the selected lags \((d_y,d_c)\) and the decision threshold \(\tau\).

\begin{table*}[ht]
\centering
\scriptsize
\setlength{\tabcolsep}{4pt}
\renewcommand{\arraystretch}{1.15}

\begin{tabular}{P G C C C}
\toprule
\textbf{Classifier} & \textbf{Hyperparameters grid} & \textbf{ITW} & \textbf{ITW-d1} & \textbf{EXW} \\
\midrule
XGBoost &
\begin{tabular}[t]{@{}l c l@{}}
\code{learning\_rate} & = & \code{[0.01, 0.05, 0.1]} \\
\code{max\_depth} & = & \code{\{3, 5, 10\}} \\
\code{n\_estimators} & = & \code{[50, 100, 200]} \\
\code{reg\_alpha} & = & \code{[0, 0.5, 1]} \\
\code{reg\_lambda} & = & \code{[1, 3, 10]} \\
\code{scale\_pos\_weight} & = & \code{[1, 3.7, 5, 10, 20.88]}
\end{tabular}
&
\code{lr=0.1, n=100, depth=5, ra=0.5, rl=1.0, spw=3.7; H=7, dy=14, dc=14, $\tau$=0.5}
&
\code{lr=0.1, n=100, depth=3, ra=0.5, rl=3.0, spw=20.88; H=7, dy=28, dc=28, $\tau$=0.4}
&
\code{lr=0.1, n=50, depth=3, ra=0.0, rl=3.0, spw=1; H=7, dy=14, dc=28, $\tau$=0.2} \\
\addlinespace[3pt]

Logistic regression &
\begin{tabular}[t]{@{}l c l@{}}
\code{C} & = & \code{[0.01, 0.1, 1, 5, 10]} \\
\code{penalty} & = & \code{\{l1, l2\}} \\
\code{class\_weight} & = & \code{\{None, balanced\}}
\end{tabular}
&
\code{C=0.1, pen=l1, cw=None; H=7, dy=28, dc=14, $\tau$=0.5}
&
\code{C=0.1, pen=l2, cw=None; H=7, dy=14, dc=14, $\tau$=0.25}
&
\code{C=5, pen=l1, cw=None; H=7, dy=28, dc=28, $\tau$=0.1} \\
\addlinespace[3pt]

LightGBM &
\begin{tabular}[t]{@{}l c l@{}}
\code{learning\_rate} & = & \code{[0.01, 0.05, 0.1]} \\
\code{n\_estimators} & = & \code{[50, 100, 200]} \\
\code{reg\_alpha} & = & \code{[0, 0.5, 1]} \\
\code{reg\_lambda} & = & \code{[1, 3, 10]} \\
\code{scale\_pos\_weight} & = & \code{[1,3.7,10, 20.88]}
\end{tabular}
&
\code{lr=0.05, n=50, ra=0.5, rl=3.0, spw=3.7; H=7, dy=14, dc=14, $\tau$=0.45}
&
\code{lr=0.05, n=50, ra=0.5, rl=3.0, spw=20.88; H=7, dy=14, dc=28, $\tau$=0.2}
&
\code{lr=0.05, n=100, ra=0.5, rl=1.0, spw=10; H=7, dy=14, dc=14, $\tau$=0.35} \\
\addlinespace[3pt]

Decision tree &
\begin{tabular}[t]{@{}l c l@{}}
\code{criterion} & = & \code{\{gini, entropy\}} \\
\code{max\_depth} & = & \code{\{5, 10, 15, None\}} \\
\code{min\_samples\_split} & = & \code{[2, 5, 10]} \\
\code{min\_samples\_leaf} & = & \code{[1, 3, 5]} \\
\code{class\_weight} & = & \code{\{None, balanced\}}
\end{tabular}
&
\code{crit=entropy, depth=None, split=2, leaf=5, cw=None; H=7, dy=28, dc=14, $\tau$=0.6}
&
\code{crit=entropy, depth=15, split=2, leaf=5, cw=balanced; H=7, dy=14, dc=14, $\tau$=0.8}
&
\code{crit=entropy, depth=10, split=2, leaf=5, cw=balanced; H=7, dy=14, dc=14, $\tau$=0.75} \\
\addlinespace[3pt]

MLP &
\begin{tabular}[t]{@{}l c l@{}}
\code{hid\_layer\_sizes} & = & \code{\{4, 16 32,64,128,64\}} \\
\code{alpha} & = & \code{[1e-4, 1e-3, 1e-2]} \\
\code{batch\_size} & = & \code{\{32, 64\}} \\
\code{max\_iter} & = & \code{\{100, 200\}}
\end{tabular}
&
\code{hls=(4,), alpha=1e-3, bs=32, iters=200; H=7, dy=14, dc=28, $\tau$=0.75}
&
\code{hls=(4,), alpha=1e-3, bs=32, iters=100; H=7, dy=28, dc=28, $\tau$=0.3}
&
\code{hls=(128,64), alpha=1e-3, bs=64, iters=200; H=7, dy=14, dc=14, $\tau$=0.25} \\
\addlinespace[3pt]

HistGradientBoosting &
\begin{tabular}[t]{@{}l c l@{}}
\code{learning\_rate} & = & \code{[0.01, 0.05, 0.1]} \\
\code{max\_depth} & = & \code{\{3, 5, 6, 10, None\}} \\
\code{max\_iter} & = & \code{\{50, 100, 200\}} \\
\code{l2\_regularization} & = & \code{[0.0, 0.01, 0.1]}\\
\code{n\_iters} & = & \code{[10,50,100]}\\
\end{tabular}
&
\code{lr=0.1, depth=6, iters=50, l2=0.01; H=7, dy=28, dc=14, $\tau$=0.25}
&
\code{lr=0.1, depth=6, iters=100, l2=0.01; H=7, dy=28, dc=14, $\tau$=0.15}
&
\code{lr=0.1, depth=None, iters=100, l2=0.01; H=7, dy=14, dc=28, $\tau$=0.3} \\
\addlinespace[3pt]

Random forest &
\begin{tabular}[t]{@{}l c l@{}}
\code{n\_estimators} & = & \code{[50, 100, 200]} \\
\code{max\_depth} & = & \code{\{5, 10, 20, None\}} \\
\code{min\_samples\_split} & = & \code{[2, 5, 10]} \\
\code{min\_samples\_leaf} & = & \code{[1, 3, 5]} \\
\code{class\_weight} & = & \code{\{None, balanced\}}
\end{tabular}
&
\code{n=50, depth=None, split=2, leaf=3, cw=balanced; H=7, dy=14, dc=14, $\tau$=0.4}
&
\code{n=50, depth=None, split=5, leaf=3, cw=balanced; H=7, dy=14, dc=28, $\tau$=0.15}
&
\code{n=50, depth=5, split=2, leaf=5, cw=balanced; H=7, dy=14, dc=14, $\tau$=0.45} \\
\bottomrule
\end{tabular}

\clearpage

\caption{Best hyperparameters identified by grid search for machine learning models.}
\label{tab:hp_results_ml}
\end{table*}

\newpage

Table~\ref{tab:hp_results_deep} lists the grids and selected settings for sequence models at \(H{=}7\). Architectures are implemented in \texttt{PyTorch} via \texttt{Darts}. Since binary heads are not provided natively, we added a sigmoid output layer and train with a binary cross-entropy loss so that models output calibrated class-1 probabilities. The sequence length \(\texttt{seq\_len}\) is tuned over \(\{14,21,28\}\). For each dataset, the table reports the chosen architecture settings along with the batch size (\texttt{bs}), number of epochs (\texttt{ep}), and the selected threshold \(\tau\).
\begin{table*}[ht]
\centering
\scriptsize
\setlength{\tabcolsep}{4pt}
\renewcommand{\arraystretch}{1.15}

\begin{tabular}{P G C C C}
\toprule
\textbf{Classifier} & \textbf{Hyperparameters grid} & \textbf{ITW} & \textbf{ITW-d1} & \textbf{EXW} \\
\midrule
LSTM--MIMO &
\begin{tabular}[t]{@{}l c l@{}}
\code{seq\_len} & = & \code{[14, 21, 28]} \\
\code{hidden\_dim} & = & \code{[64, 128]} \\
\code{n\_rnn\_layers} & = & \code{[1, 2]} \\
\code{dropout} & = & \code{[0.0, 0.2, 0.5]} \\
\code{activation} & = & \code{\{relu,tanh\}}
\end{tabular}
&
\code{seq=14; H=7; act=relu, hid=128, layers=1, dr=0.5; bs=64, ep=30; $\tau$=0.2}
&
\code{seq=14; H=7; act=relu, hid=128, layers=2, dr=0.0; bs=128, ep=30; $\tau$=0.1}
&
\code{seq=28; H=7; act=relu, hid=64, layers=2, dr=0.5; bs=128, ep=75; $\tau$=0.1} \\
\addlinespace[3pt]

LSTM--Seq2Seq &
\begin{tabular}[t]{@{}l c l@{}}
\code{seq\_len} & = & \code{[7, 14,28]} \\
\code{hidden\_dim} & = & \code{[64, 128]} \\
\code{n\_layers} & = & \code{[1, 2, 3]} \\
\code{dropout} & = & \code{[0.0, 0.2, 0.5]} \\
\code{activation} & = & \code{\{relu,tanh\}}\\
\end{tabular}
&
\code{seq=14; H=7; hid=64,act=relu, layers=2, dr=0.5; bs=64, ep=50; $\tau$=0.95}
&
\code{seq=28; H=7; hid=64,act=relu, layers=1, dr=0.5; bs=64, ep=30; $\tau$=0.2}
&
\code{seq=14; H=7; hid=128, layers=2, dr=0.5; bs=32, ep=30; $\tau$=0.85} \\
\addlinespace[3pt]

TCN &
\begin{tabular}[t]{@{}l c l@{}}
\code{seq\_len} & = & \code{[7, 14,28]} \\
\code{num\_filters} & = & \code{[16, 32, 64]} \\
\code{kernel\_size} & = & \code{[3, 5]} \\
\code{dilation\_base} & = & \code{[2, 3]} \\
\code{dropout} & = & \code{[0.0, 0.2, 0.5]} \\
\code{weight\_norm} & = & \code{\{True, False\}}
\end{tabular}
&
\code{seq=14; filt=64, k=3, dil=3, dr=0.5, wn=False; bs=64, ep=50; $\tau$=0.55}
&
\code{seq=28; filt=64, k=5, dil=3, dr=0.2, wn=False; bs=64, ep=50; $\tau$=0.65}
&
\code{seq=14;  filt=64, k=5, dil=3, dr=0.5, wn=True; bs=32, ep=50; $\tau$=0.7} \\
\addlinespace[3pt]

TFT &
\begin{tabular}[t]{@{}l c l@{}}
\code{seq\_len} & = & \code{[7, 14,28]} \\
\code{hidden\_size} & = & \code{[32, 64]} \\
\code{attn\_heads} & = & \code{[2, 4, 8]} \\
\code{dropout} & = & \code{[0.0, 0.2, 0.5]}
\end{tabular}
&
\code{seq=14; hid=32, heads=4, dr=0.1; bs=64, ep=30; $\tau$=0.50}
&
\code{seq=14;  hid=64, heads=4, dr=0.1; bs=64, ep=30; $\tau$=0.40}
&
\code{seq=28;  hid=32, heads=4, dr=0.1; bs=64, ep=30; $\tau$=0.30} \\
\bottomrule
\end{tabular}

\caption{Best hyperparameters identified by grid search for deep learning models.}
\label{tab:hp_results_deep}
\end{table*}

\twocolumn
\bibliographystyle{elsarticle-harv} 
\bibliography{bibliography}





\end{document}